%% file: main.tex
\documentclass[runningheads]{llncs}

\usepackage{PRIMEarxiv}

\usepackage{tabularx} 
\usepackage{float} 
\usepackage{caption} 
\usepackage[justification=centering]{caption} 
\usepackage{listings}
\usepackage{xcolor}
\usepackage{multirow}
\usepackage{cite}
\usepackage[a4paper,pagebackref,hyperindex=true]{hyperref}

\usepackage{enumerate}

\usepackage[automake,acronym,toc,style=listdotted,shortcuts]{glossaries-extra}
\glsdisablehyper
\setabbreviationstyle[acronym]{long-short}
\setabbreviationstyle[abbreviation]{short}
\makeglossaries
\usepackage{siunitx}

\usepackage{graphicx}

\usepackage{longtable}
\usepackage{booktabs}

\begin{document}

\pagestyle{headings}
\mainmatter
\title{Image-based Detection of Surface Defects in Concrete during Construction}

\author{
   Jan Dominik Kuhnke \\
   Technische Universität Berlin \\
   Germany, Berlin \\
   \And
  Monika Kwiatkowski \\
  Computer Vision \& Remote Sensing \\
  Technische Universität Berlin \\
   Germany, Berlin \\
     \And
  Olaf Hellwich \\
  Computer Vision \& Remote Sensing \\
  Technische Universität Berlin \\
   Germany, Berlin \\
}

\graphicspath{{.}{images/}}

\maketitle

\begin{abstract}

Defects increase the cost and duration of construction projects as they require significant inspection and documentation efforts. Automating defect detection could significantly reduce these efforts. This work focuses on detecting honeycombs, a substantial defect in concrete structures that may affect structural integrity.

We compared honeycomb images scraped from the web with images obtained from real construction inspections. We found that web images do not capture the complete variance found in real-case scenarios and that there is still a lack of data in this domain. Our dataset is therefore freely available for further research.

A Mask R-CNN and EfficientNet-B0 were trained for honeycomb detection. The Mask R-CNN model allows detecting honeycombs based on instance segmentation, whereas the EfficientNet-B0 model allows a patch-based classification.
Our experiments demonstrate that both approaches are suitable for solving and automating honeycomb detection. In the future, this solution can be incorporated into defect documentation systems. \\

\begin{figure}[h]
\begin{center}
\includegraphics[width=50mm]{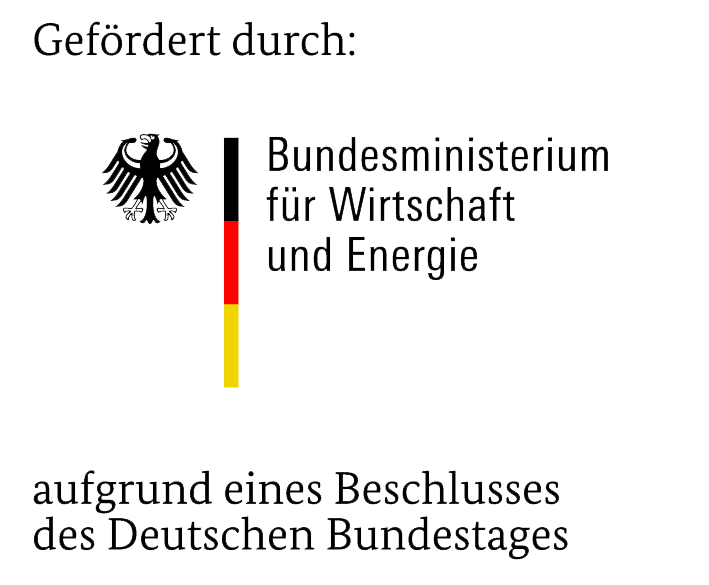} \\
\end{center}
\end{figure}

This work was carried out as part of the research project "Smart Design and Construction Through Artificial Intelligence" (SDaC). SDaC is funded by the German Federal Ministry of Economics and Technology (BMWi).

\end{abstract}

\include{Chapter0_abbr}

\hypersetup{hidelinks}

\input{Chapter_Introduction}

\input{Chapter_Methodology}
\input{Chapter_Results}
\input{Chapter_Conclusion}


\bibliographystyle{ieeetr}
\bibliography{Thesis}


\end{document}

%% file: Chapter0_abbr.tex

\newacronym{cnn}{CNN}{convolutional neural network}
\newacronym{gdp}{GDP}{gross domestic product}
\newacronym{iou}{IoU}{intersection of union}
\newacronym{cvat}{CVAT}{computer vision annotation tool}
\newacronym{dbv}{DBV}{Deutscher Beton- und Bautechnik-Verein e.V. (English: German concrete and construction technology association e.V.)}
\newacronym{xai}{XAI}{explainable artificial intelligence}
\newacronym{ai}{AI}{artificial intelligence}
\newacronym{cv}{CV}{computer vision}
\newacronym{ml}{ML}{machine learning}
\newacronym{gan}{GAN}{generative adversarial network}
\newacronym{vcp}{VCP}{vitrified clay pipe}
\newacronym{ap}{AP}{average precision}
\newacronym{ar}{AR}{average recall}
\newacronym{map}{mAP}{mean average precision}
\newacronym{iid}{i.i.d.}{independent and identically distributed}
\newacronym{rpn}{RPN}{region proposal network}
\newacronym{bgb}{BGB}{Bürgerliches Gesetzbuch (English: German Civil Code)}
\newacronym{vob}{VOB}{Vergabe- und Vertragsordnung für Bauleistungen (English: German Construction Contract Procedures)}
\newacronym{mrcnn}{Mask R-CNN}{Mask R-CNN} \glsunset{mrcnn}
\newacronym{sdac}{SDaC}{smart design and construction}
\newacronym{bmwi}{BMWi}{Bundesministerium für Wirtschaft und Klimaschutz (English: Federal Ministry for Economic Affairs and Climate Action)}
\newacronym{cam}{CAM}{class activation mapping}
\newacronym{uav}{UAV}{unmanned aerial vehicle}
\newacronym{ugv}{UGV}{unmanned ground vehicle}
\newacronym{fcn}{FCN}{fully convolutional network}
\newacronym{hicc}{HiCC}{honeycombs in concrete classification}
\newacronym{hicis}{HiCIS}{honeycombs in concrete instance segmentation}
\newacronym{auc}{AUC}{area under curve}
\newacronym{flop}{FLOP}{floating point operation}
\newacronym{cdc}{CDC}{concrete damage classification}
\newacronym{cdc-bhc}{cdc-bhc}{concrete damage classification - binary honeycomb classification}
\newacronym{ccifc}{CCIfC}{concrete crack images for classification}
\newacronym{bim}{BIM}{building information model}

\newacronym{gradCam}{Grad-CAM}{gradient-weighted class activation mapping}
\newacronym{rebar}{rebar}{reinforcing bar}
\newacronym{eg}{e.g.}{exempli gratia}
\newacronym{ie}{i.e.}{id est}
\newacronym{etal}{et al.}{et alii, et aliae, et aelia}
\newacronym{vs}{vs.}{versus}


%% file: Chapter_Introduction.tex

\section{Introduction}

\paragraph{} 
Construction defects are costly for the economy.
The cost of defect elimination is between 2\% and 12.4\% of the total cost of construction \cite{lundkvist2014proactive} and much time and effort is required to inspect construction sites and document defects \cite{nguyen2015smart}.

Automating the inspection of construction projects would free up resources and may even enable more frequent inspections, leading to more efficient construction projects.
The progress in CV and ML may enable the complete automation of this process in the future. 

Although deep learning is applied to many different fields, research into image-based defect detection using deep learning is still limited in the construction industry, despite its large size, and focuses on security, progress, and productivity. 

In contrast, there appear to be relatively few publications on methods utilized for object detection in quality assurance in construction. So far, the research into detecting defects has been mainly limited to defects occurring in the maintenance phase of infrastructure facilities such as roads, bridges, and sewer systems. \cite{xu2020computer}

This work focuses on the detection of honeycombs, which are large surface voids in concrete, that often contain visible pebbles as the lack of cement reveals the gravel. Honeycombs may expose reinforcements, \acs{ie} \glspl{rebar}, leading to erosion, and impair the water impermeability and static strength of the concrete.

\section{Related Work}
Most research into honeycomb detection uses a variety of sensor data, but not images.
For example, Ismail and Ong (2012) \cite{ismail2012honeycomb} used mode shapes to detect honeycombs in reinforced concrete beams. Vibration is induced into the concrete beam, and the displacement caused is measured at specific locations on the beams, describing the behavior of an object under dynamic load. Furthermore, V\"olker and Shokouhi \cite{volker2015clustering} developed a multi-sensor clustering-based method for honeycomb detection, using impact-echo, ultrasound, and ground-penetrating radar data.

To the author's best knowledge, the following is the only work using regular camera images and applying \gls{ml} for honeycomb detection and is used as a baseline for this thesis.
Hung \acs{etal}\ \cite{hung2019surface} showed that \glspl{cnn} could classify concrete images with a precision of 93\% and recall of 93\% into honeycomb, crack, moss, blistering, and normal classes. When their dataset, \gls{cdc}, was published, it was limited in size and was scraped from the internet, introducing a bias, as demonstrated in Section \ref{sec:resClassificaiton}. Regarding honeycombs, for example, the images scraped from the internet are often explanatory illustrations and show the most easily identifiable honeycombs. Furthermore, the domain from which the pictures are drawn is limited to images of concrete, leading to the expectation of a high false positive rate when applied to realistic defect images. This cannot, therefore, prove the applicability of \glspl{cnn} for honeycomb detection for images of general defects as collected in a defect documentation system. Nevertheless, this demonstrates that \glspl{cnn} may be applicable for the detection of honeycombs in concrete, assuming the images depict only concrete.

In conclusion, research into defect detection focuses on maintenance defects and often uses images which are not created by potential users. In recent research, \Gls{cnn}-based approaches dominated the field, including classification and object detection. As a result, this work evaluates both approaches and explores differences between an existing dataset of honeycomb images and a realistic dataset of images taken by construction site inspectors.

%% file: Chapter_Methodology.tex
\section{Methodology}
While multiple datasets of defects in concrete structures exist, they either focus on different defects or contain images scraped from the internet. Hung \acs{etal}\ \cite{hung2019surface} published the most relevant dataset used for honeycomb detection, but this was only published after data augmentation, increasing relabeling effort. 
Two datasets are collected. First, \emph{Metis Systems AG} provided a set of honeycomb images. Second, similarly to Hung \acs{etal}\ \cite{hung2019surface}, a dataset was scraped from the internet, providing a baseline to similar scraped datasets from research and enabling comparison with a realistic dataset.

\subsection{Data origins}
In the context of the research project \emph{SDaC}, \emph{Metis Systems AG} provided access to the defect images documented in their proprietary software \emph{überbau}. Inspectors document defects in \emph{überbau}, assigning each defect a title, optionally any number of images, and more attributes. These photos are often taken with a smartphone, at a variety of camera angles and lighting conditions may range from brightly sunlit to dark and lit only by a flash. The defects were accessed via an internal API and filtered for the key word "honeycomb", resulting in a total of 780 images. These images were further manually classified into 191 \emph{honeycomb}, and 539 \emph{other}. The honeycomb class contains images of easily identifiable honeycombs. In contrast, \emph{other} contains images of unrelated scenery and honeycombs, that are too difficult to differentiate for initial research and would have required more resources.

To the author's knowledge, this dataset is the most extensive public collection of honeycombs in existence and is the only one comprised of images that are neither scraped from the internet nor taken especially for research. 
 
Furthermore, all images are made public with permission of \emph{Metis Systems AG} and may be used for further research. While our definition for honeycombs may not fit that of other researchers, the raw images are also published and increase the number of publicly available photos of concrete structures with honeycombs, pores, etc., caused by errors during construction rather than deterioration.

The second dataset was obtained using a google image search similarly to Hung \acs{etal}\ \cite{hung2019surface}, who collected images for four classes. However, since only the class of honeycombs is relevant for this work, only the keyword phrases \emph{honeycomb concrete} and \emph{honeycomb on concrete surface} were used, and the number of downloaded images was increased from 50 to 100 compared to Hung \acs{etal}\ \cite{hung2019surface}. The first 100 images matching the keyword phrases were downloaded and images containing no honeycombs, those with watermarks obstructing the image, duplicates, and images of low quality were removed. The filtered images were then combined and duplicates were removed a second time resulting in 56 images depicting honeycombs. 

However, the images scraped from the internet are primarily used as examples of honeycombs by different websites, introducing a bias for very clear and large honeycombs. As a result most images contain very noticeable pebble-like structures.

\subsection{Datasets}\label{sec:datasets}

Since honeycombs do not describe an object per se, the determination of its outline is a challenge in itself. In contrast, most pores have clear circular outlines as highlighted by previous work \cite{liu2017image, zhu2008detecting, zhu2010machine, yoshitake2018image,nakabayash2020automatic}.
We use two approaches to labeling our data. First, we create instance segmentation masks. Secondly, we use simple classification labels.
We apply our labeling techniques to both our datasets.

\subsubsection{Honeycombs in concrete instance segmentation}\label{sec:hicis}
The conventional definition of a honeycomb is a surface void exceeding a certain diameter. This definition can not be applied here, since estimating the size of surface voids is impossible due to the unknown scale of the images. Some works address this issue by defining a fixed area \cite{nakabayash2020automatic} or controlling the image capture process \cite{yoshitake2018image}, resulting in a fixed scale of the surfaces displayed in the images. However, our solution avoids these biases and enables detection at any scale. 
We use the following definition as a labeling criterion: a honeycomb is a surface void in concrete with at least one partially visible pebble.
Finally, the \gls{hicis} datasets were created by labeling the images with instance segmentation masks according to the aforementioned honeycomb definition. 

The datasets were split into train, validation, and test sets of 60\%, 20\%, and 20\%, respectively, creating the two datasets \emph{\gls{hicis} metis} and \emph{\gls{hicis} web} with three subsets each.

\subsubsection{Honeycombs in concrete classification}\label{sec:hicc}

In addition to our segmentation labels, we create several classification datasets. First, we use the \gls{cdc} dataset by Hung \acs{etal}\ \cite{hung2019surface}. The dataset was converted from multi to binary classification for honeycombs by sorting all non-honeycomb images into a single class. The resulting dataset is called \gls{cdc-bhc}.

Additionally, the classification datasets \gls{hicc} were created from our \gls{hicis} segmentation datasets.
Square patches of $224\times224$ pixels were generated from the \gls{hicis} dataset by cropping the images, then calculating the area of the instance segmentation mask in the cropped image and applying a threshold to binary classification labels.

The crop was moved over the image by either a slide of 112 pixels or 224 pixels, creating two datasets for each \gls{hicis} dataset.
The datasets created with a slide of 112 pixels contain each pixel up to four times in different patches while the others contain each pixel exactly once. Keeping the train, validation, and test splits from \gls{hicis}, assured that a specific honeycomb does not appear in different subsets. Table \ref{tab:overviewHiccDatasets} summarizes the datasets. The generated datasets originating from the \gls{hicis} dataset follow this naming convention: \emph{HiCC/\{origin\}-s\{stride\_size\}-p\{patch\_size\}}.
\begin{table}[H]
	\footnotesize
	\centering
	\begin{tabular}{!{\extracolsep{4pt}}rrrrrrrr}
		& & \multicolumn{2}{c}{\textbf{train}} & \multicolumn{2}{c}{\textbf{validation}} & \multicolumn{2}{c}{\textbf{test}}\\
		\cline{3-4}\cline{5-6}\cline{7-8}
		\textbf{origin} & \textbf{dataset name} & \textbf{true} & \textbf{false} & \textbf{true} & \textbf{false} & \textbf{true} & \textbf{false}\\
		\toprule
		
		\gls{cdc} \cite{hung2019surface} & \gls{cdc-bhc}-224 	& 840 & 3360 & 0 & 0 & 210 & 840\\
		\gls{hicis}-metis & \gls{hicc}-metis-s124-p224 		    & 10480 & 64359 & 3684 & 25014 & 4281 & 24700\\
		\gls{hicis}-metis & \gls{hicc}-metis-s224-p224 			& 2676 & 16976 & 936 & 6571 & 1080 & 6498 \\
		\gls{hicis}-web & \gls{hicc}-web-s124-p224 				& 573  & 823 & 156 & 28 & 132 & 20\\
		\gls{hicis}-web & \gls{hicc}-web-s224-p224 				& 161 & 231 & 48 & 8 & 44 & 5\\
	\end{tabular}
	\caption{Overview of classification datasets}\label{tab:overviewHiccDatasets}
\end{table}
The procedure generated a relatively large number of samples; more than three times larger than the original \gls{cdc} dataset. 

\subsection{Transfer-learning}
CNNs often require a large amount of data and, even on modern systems, training time can span multiple days. Therefore, pre-trained models were used, reducing training time and improving the generalization of the models, as demonstrated by Özgenel and Sorguç \cite{ozgenel2018performance} for crack detection in concrete. 

\subsubsection{Mask R-CNN with ResNet101 Backbone for Instance Segmentation}

\gls{mrcnn} architecture is used for instance segmentation. \Gls{mrcnn} achieved state-of-the-art performance for COCO at the time of publication by He \acs{etal}\ \cite{he2018mask} in 2017.

We used \emph{ResNet101} as a backbone.

A warmup phase starting from a learning rate of $5e-6$ and reaching $5e-3$ at the 100th epoch was used. The learning rate was then constant until the 2000th iteration onwards, when the learning rate was halved every 250 iterations except for the models trained on only \gls{hicis} \emph{metis}, for which halving started at the 1000th iteration. 512 regions of interest were generated per image. Since the GPU memory was limited, images were resized to 1024px x 1024px, and a batch size of 2 images per iteration was used. The models were trained for a total of 6000 iterations.

\subsubsection{EfficientNet for classification}\label{sec:enb-cdc}
The number of \Gls{cnn}-based classification models developed since its inception in 1995 is vast \cite{li2021survey}. 
Hung \acs{etal}\ \cite{hung2019surface} used VGG19 \cite{simonyan2014very}, InceptionV3 \cite{szegedy2016rethinking} and InceptionResNetV2 \cite{szegedy2017inception} for their successful training of classifying surface damages in concrete. EfficientNet architectures are prevalent because of their performance to parameter ratio \cite{tan2019efficientnet}. 

EfficientNet-L2, the best model of the EfficientNet architectures, achieves 90.2\% top 1 accuracy on Imagenet.

Since Hung \acs{etal}\ \cite{hung2019surface} used an input image size of 227x227p, the closest matching EfficientNet model, EfficientNet-B0, is used. 

The application of transfer learning consisted of three stages.
First, only the output layer was trained. Second, in addition to the output layer, the last block of the EfficientNet-B0 was also trained to increase the number of trainable parameters without losing the low-level feature extractors in the early blocks. Third, the model was trained in its entirety. All training stages used Adam \cite{kingma2014adam} for optimization. 

To demonstrate the viability and effectiveness of EfficientNet-B0, the model was trained on the \gls{cdc} dataset by Hung \acs{etal}, without adding any additional augmentations since \gls{cdc} is already augmented. Table \ref{tab:resCdcTrainParams} displays the training parameters.
\begin{table}[H]
	\centering
	\begin{tabular}{ccccc}
		\textbf{stage} & \textbf{epoch} & \textbf{initial learning rate} & \textbf{beta1} & \textbf{beta2} \\
		\toprule
		1 & 1 & 1e-2 & 0.9 & 0.9 \\
		2 & 1 & 1e-5 & 0.9 & 0.9 \\
		3 & 8 & 1e-8 & 0.9 & 0.9 \\
	\end{tabular}
	\caption{Parameters for training EfficientNet-B0 using Adam on the CDC dataset}\label{tab:resCdcTrainParams}
\end{table}

Table \ref{tab:resHicTrainParams} describes the training stages omitting the $beta1$ and $beta2$ values of the Adam optimizer since they all are set to $0.9$. In addition to randomly changing contrast, saturation, and brightness, the JPG quality of the input images was randomly set between 50 to 100.

\begin{table}[H]
	\centering
	\begin{tabular}{!{\extracolsep{4pt}}rcccccc}
		 & \multicolumn{2}{c}{\textbf{stage 1}} & \multicolumn{2}{c}{\textbf{stage 2}} & \multicolumn{2}{c}{\textbf{stage 3}} \\
		 \cline{2-3}\cline{3-4}\cline{5-6}
		\textbf{training dataset} & \textbf{epochs} & \textbf{lr} & \textbf{epochs} & \textbf{lr} & \textbf{epochs} & \textbf{lr} \\
		\toprule
		
		\gls{cdc-bhc} & 1 & 1e-1 & 1 &  1e-4& 4& 1e-7 \\
		\gls{hicc}-metis-s112-p224 & 1 & 1e-1 & 1& 1e-4& 4& 1e-7\\
		\gls{hicc}-metis-s224-p224 & 1 & 1e-2 & 1& 1e-5 & 4 & 1e-8 \\
		\gls{hicc}-web-s112-p224 & 1 & 1e-1 & 1 & 1e-4&  4& 1e-7 \\
		\gls{hicc}-web-s224-p224 & 1 & 1e-2 & 1& 1e-5 & 4& 1e-8 \\
		concat-s112-p224 & 1 & 1e-2 & 1 & 1e-5 & 1 & 1e-8 \\
		concat-s224-p224 & 1 & 1e-2 & 1& 1e-5 & 4& 1e-8 \\
	\end{tabular}
	\caption{Parameters for training EfficientNet-B0 using Adam with $beta1=beta2=0.9$ on binary honeycomb datasets}\label{tab:resHicTrainParams}
\end{table}

\emph{Concat-s112-p224} describes the combination of \emph{metis-s112-p224} and \emph{web-s112-p224} and \emph{concat-s112-p224} including the non-sliding versions, meaning a slide of 224 pixels and a patch size of 224 pixels. 

%% file: Chapter_Results.tex
\section{Results and Discussion}
\label{chap:results}

\subsection{EfficientNet-B0 for Concrete Damage Classification}
EfficientNet-B0 achieved better performance than all models by Hung \acs{etal}\ \cite{hung2019surface}, technically reaching state of the art for the \gls{cdc} dataset.
After ten training epochs, EfficientNet-B0 achieves an accuracy of 96.95\%, a precision of 97.32\%, and a recall of 96.76\% on the \gls{cdc} dataset. The precision and recall are higher for each class and on average than in the former best-performing model, InceptionResnetV2, as shown by Table \ref{tab:compareBlub}. Furthermore, EfficientNet-B0's accuracy of 96.29\%  is statistically significantly higher than VGG19's 92.29\%, InceptionV3's 90.57\%, and InceptionResnetV2's 92.57\%. 
In conclusion, these results demonstrate that EfficientNet-B0 should be able to reach satisfactory performance on realistic datasets if the images scraped from the web sufficiently resemble honeycombs.

\begin{table}[H]
\centering
		\begin{tabular}{rcccc}
			\textbf{class} & \textbf{precision} & \textbf{recall} & \textbf{f1-score} & 
			\textbf{support} \\
			\toprule
			
			normal & 0.96 & 0.97 & 0.96 & 210 \\
			cracked & 0.95 & 0.97 & 0.96 & 210 \\
			blistering & 0.98 & 0.96 & 0.97 & 210 \\
			honeycomb & 0.96 & 0.98 & 0.97 & 210 \\
			moss & 1.00 & 0.97 & 0.98 & 210 \\
			average & 0.97 & 0.97 & 0.97 & 1050 \\
			
		\end{tabular}\\
			$(a)$ Our finetuned EfficientNet-B0
\end{table}		
\begin{table}[H]
\centering
		\begin{tabular}{rcccc}
			\textbf{class} & \textbf{precision} & \textbf{recall} & \textbf{f1-score} & 
			\textbf{support} \\
			\toprule
			
			normal & 0.91 & 0.93 & 0.92 & 210 \\
			cracked & 0.90 & 0.90 & 0.90 & 210 \\
			blistering & 0.94 & 0.91 & 0.93 & 210 \\
			honeycomb & 0.89 & 0.97 & 0.93 & 210 \\
			moss & 0.99 & 0.91 & 0.95 & 210 \\
			
			avg & 0.93 & 0.93 & 0.93 & 1050 \\
		\end{tabular}
		\\
	  $(b)$ InceptionResnetV2 by Hung \acs{etal}\ \cite{hung2019surface} \\
	\caption{Comparison of performance on the CDC dataset of our finetuned EfficientNet-B0 vs. InceptionResnetV2 by Hung et al. \cite{hung2019surface}}\label{tab:compareBlub}
\end{table}

\subsection{Classification Results}\label{sec:resClassificaiton}
The training of EfficientNet-B0 showed clear differences between the datasets in \gls{hicc}. 

All models performed the best on their own test sets, considering \gls{ap} and \gls{ar}.
Each model achieved high performance on the \gls{hicc} \emph{web} datasets.
Table \ref{tab:enb0Metrics} displays the metrics of each model on the different test sets.

\begin{table}[H]
	\centering
	\begin{tabular}{!{\extracolsep{4pt}}rccccr}
		
		\textbf{test set} & \textbf{precision} & \textbf{recall} & \textbf{ap} & \textbf{ar} & \textbf{support} \\
		\toprule
		\addlinespace[1ex] & \multicolumn{5}{c}{\textbf{cdc-bhc}} \\
		\cline{2-6}
		cdc-bhc & \textbf{0.980} & \textbf{0.943} & \textbf{0.927} & \textbf{0.972} & 210 \\
		web-s112-p224 & 0.974 & 0.848 & 0.980 & 0.970 & 132 \\
		web-s224-p224 & \textbf{1.000} & 0.818 & 0.981 & 0.973 & 44 \\
		metis-s112-p224 & 0.238 & 0.242 & 0.189 & 0.220 & 4281 \\
		metis-s224-p224 & 0.226 & 0.235 & 0.181 & 0.212 & 1080 \\
		
		\addlinespace[1ex] & \multicolumn{5}{c}{\textbf{web-s224-p224}} \\
		\cline{2-6}
		cdc-bhc & 0.286 & 0.676 & 0.490 & 0.478 & 210\\
		web-s112-p224 & 0.991 & 0.879 & \textbf{0.987} & \textbf{0.978} & 132\\
		web-s224-p224 & \textbf{1.000} & 0.841 & \textbf{0.992} & \textbf{0.994} & 44\\
		metis-s112-p224 & 0.432 & 0.497 & 0.416 & 0.430 & 4281\\
		metis-s224-p224 & 0.419 & 0.490 & 0.413 & 0.425 & 1080\\
		
		\addlinespace[1ex] & \multicolumn{5}{c}{\textbf{web-s112-p224}} \\
		\cline{2-6}
		cdc-bhc & 0.295 & 0.738 & 0.347 & 0.389 & 210\\
		web-s112-p224 & 0.959 & \textbf{0.886} & 0.977 & 0.907 & 132\\
		web-s224-p224 & \textbf{1.000} & \textbf{0.864} & 0.991 & \textbf{0.994} & 44\\
		metis-s112-p224 & 0.416 & 0.404 & 0.329 & 0.342 & 4281\\
		metis-s224-p224 & 0.406 & 0.400 & 0.320 & 0.341 & 1080\\
		
		\addlinespace[1ex] & \multicolumn{5}{c}{\textbf{metis-s224-p224}}\\
		\cline{2-6}
		cdc-bhc & 0.577 & 0.448 & 0.554 & 0.550 & 210\\
		web-s112-p224 & \textbf{1.000} & 0.689 & 0.984 & 0.974 & 132\\
		web-s224-p224 & \textbf{1.000} & 0.705 & 0.991 & 0.990 & 44\\
		metis-s112-p224 & 0.621 & \textbf{0.611} & 0.640 & 0.635 & 4281\\
		metis-s224-p224 & 0.617 & \textbf{0.615} & 0.623 & 0.623 & 1080\\
		
		\addlinespace[1ex] & \multicolumn{5}{c}{\textbf{metis-s112-p224}} \\
		\cline{2-6}
		cdc-bhc & 0.644 & 0.319 & 0.444 & 0.460 & 210\\
		web-s112-p224 & 0.988 & 0.614 & 0.962 & 0.919 & 132\\
		web-s224-p224 & \textbf{1.000} & 0.591 & 0.973 & 0.963 & 44\\
		metis-s112-p224 & \textbf{0.696} & 0.568 & \textbf{0.680} & \textbf{0.677} & 4281\\
		metis-s224-p224 & \textbf{0.685} & 0.557 & \textbf{0.676} & \textbf{0.672} & 1080\\

		\addlinespace[1ex] & \multicolumn{5}{c}{\textbf{concat-s224-p224}}\\
		\cline{2-6}
		cdc-bhc & 0.714 & 0.524 & 0.654 & 0.659 & 210\\
		web-s112-p224 & 0.991 & 0.795 & 0.986 & 0.966 & 132\\
		web-s224-p224 & 1.000 & 0.773 & 0.988 & 0.983 & 44\\
		metis-s112-p224 & 0.608 & 0.597 & 0.636 & 0.634 & 4281\\
		metis-s224-p224 & 0.596 & 0.596 & 0.623 & 0.622 & 1080\\
		
		\addlinespace[1ex] & \multicolumn{5}{c}{\textbf{concat-s112-p224}}  \\
		\cline{2-6}
		cdc-bhc & 0.545 & 0.490 & 0.589 & 0.583 & 210\\
		web-s112-p224& 0.951 & 0.735 & 0.977 & 0.955 & 132\\
		web-s224-p224 & 0.969 & 0.705 & 0.983 & 0.971 & 44\\
		metis-s112-p224 & 0.637 & 0.588 & 0.633 & 0.627 & 4281\\
		metis-s224-p224 & 0.626 & 0.579 & 0.613 & 0.613 & 1080\\

	\end{tabular}
	\caption{Metrics on test sets for EfficientNet-B0 trained on different training sets}\label{tab:enb0Metrics}
\end{table}

The inclusion of the \gls{hicc} \emph{web} datasets in the \emph{metis} dataset does not improve performance on the \gls{hicc} \emph{metis} dataset. However, it improves the recall on the \gls{hicc} \emph{web} datasets. 
Overall the best model seems to be the one trained on the \gls{hicc} \emph{metis-s112-p224} dataset achieving the highest $AP$ and $AR$ for the \gls{hicc} \emph{metis} datasets and $AP$s and $ARs$ close to highest ones for the \gls{hicc} web datasets.
The high \gls{ap} and \gls{ar} achieved by this model indicate that the model achieved the most distinct separation between honeycombs and non-honeycombs. Since the \gls{ap} and \gls{ar} average their metrics at different thresholds, a high value means increased independence from picking a specific threshold. To confirm the assumption, a qualitative analysis using \acs{gradCam} is performed in Section \ref{sec:gradCramCompare}.

The higher $AR$ of the \emph{metis-s112-p224} model compared to its recall at a confidence threshold of $0.5$ is caused by the recall almost never reaching zero.

\subsubsection{Grad-CAM-based Comparison}\label{sec:gradCramCompare}
\Gls{gradCam} is a technique to highlight the regions in an image contributing the most to the prediction \cite{selvaraju2017grad}.

\Gls{gradCam} confirms that the models learned the structure of honeycombs on most datasets, although most models develop some bias for the upper left corner. The models trained on the web scraped datasets seem to learn the structure of honeycombs poorly, excluding the \emph{web-s224-p224} model, which shows a suitable \acs{gradCam}. Only the model trained on \gls{cdc-bhc} failed to classify the image correctly, and only the \emph{metis-s112-p224} model does not activate for the upper left corner.

The models whose training included \emph{metis} data handled images that are atypical to their training data satisfactorily. However, the web models whose training data did not contain such images were less confident in their predictions, although still classifying correctly. 
The higher confidence of the other models demonstrates the better separation of classes these models learned as supported by the \gls{ap} and \gls{ar} in Table \ref{tab:enb0Metrics}.

\begin{figure}[H]
	\footnotesize
	\centering
	\begin{tabular}{cccc}
	    & 0.99 & 0.94 & 1.0 \\
		\includegraphics[width=0.15\linewidth,trim={0 0 0 7mm},clip]{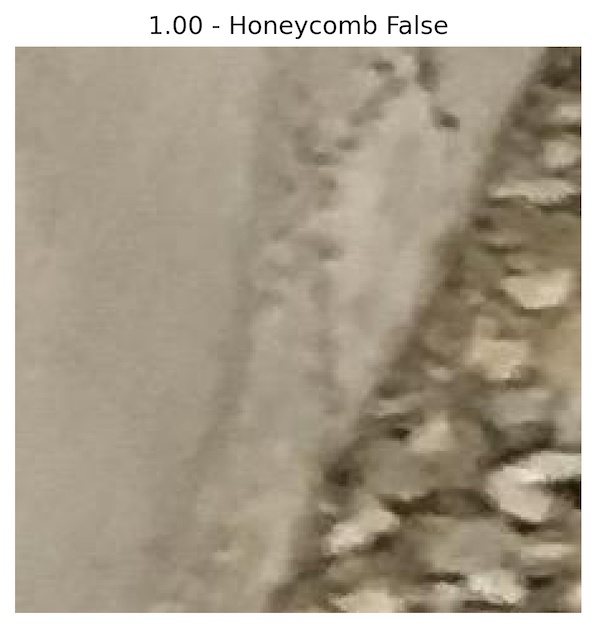}
		&
		\includegraphics[width=0.15\linewidth,trim={0 0 0 7mm},clip]{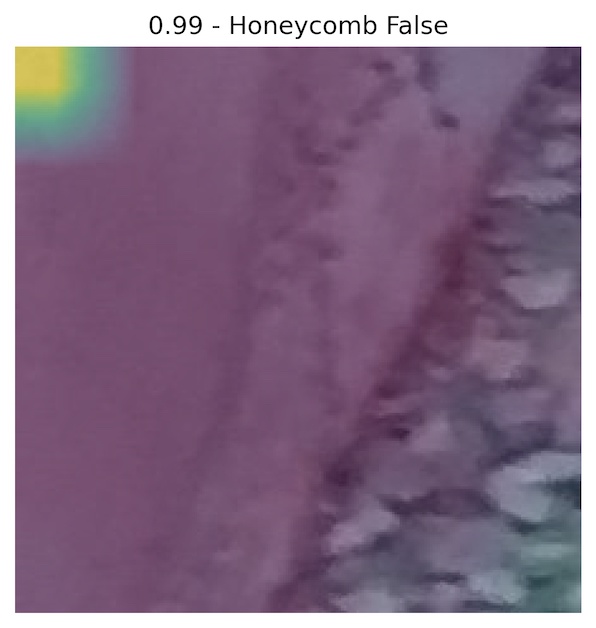}
		&
		\includegraphics[width=0.15\linewidth,,trim={0 0 0 7mm},clip]{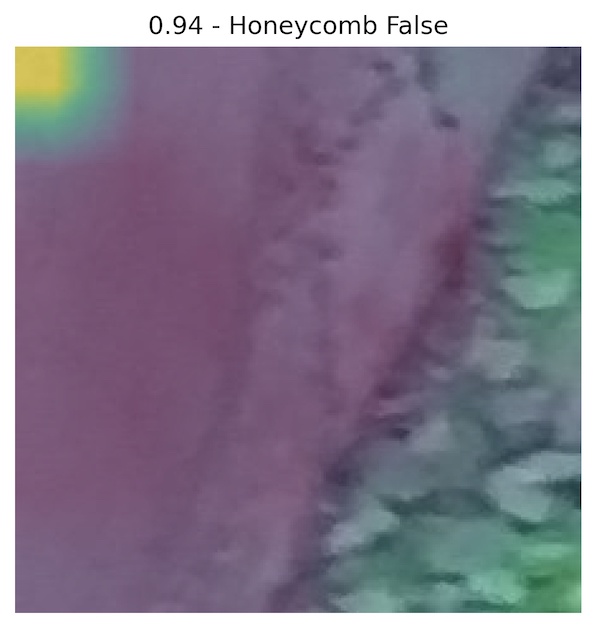}
		&
		\includegraphics[width=0.15\linewidth,trim={0 0 0 7mm},clip]{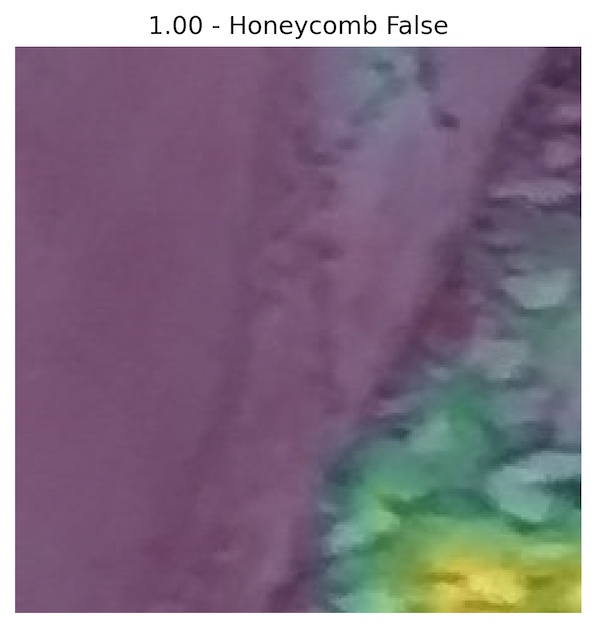}
		\\
		$(a)$ original & $(b)$ cdc-bhc & $(c)$ metis-s224-p224 & $(d)$ metis-s112-p224 \\

		\addlinespace[1ex] 0.49 & 0.99 & 1.0 & 1.0 \\
		\includegraphics[width=0.15\linewidth,trim={0 0 0 7mm},clip]{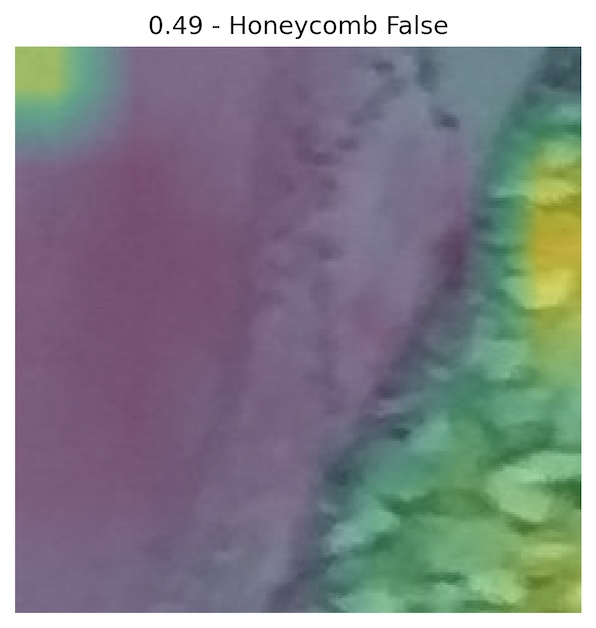}
		&
		\includegraphics[width=0.15\linewidth,trim={0 0 0 7mm},clip]{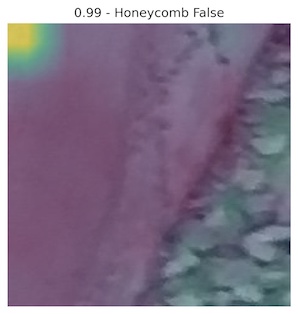}
	    &
		\includegraphics[width=0.15\linewidth,trim={0 0 0 7mm},clip]{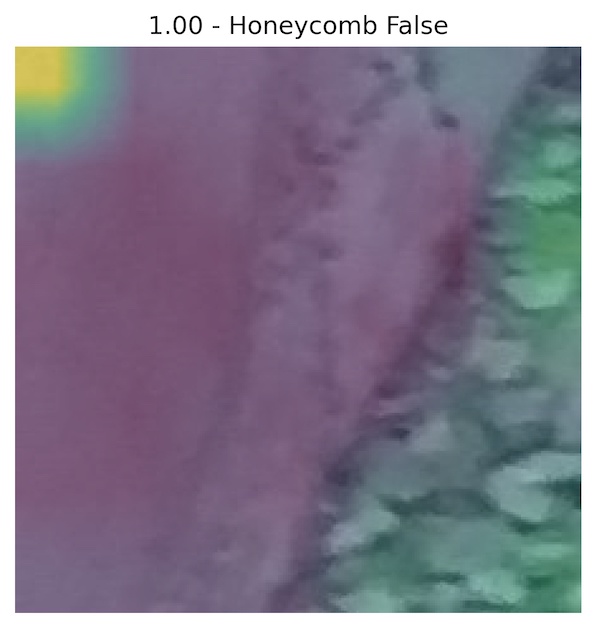}
		&
		\includegraphics[width=0.15\linewidth,trim={0 0 0 7mm},clip]{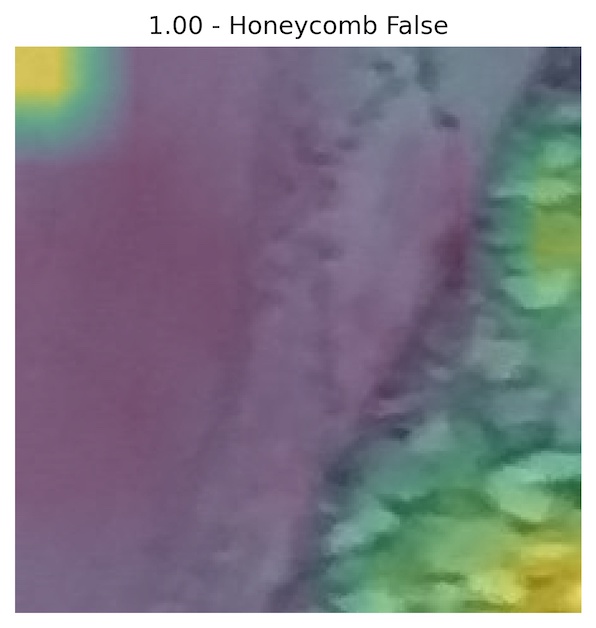}
		\\
	     $(e)$ web-s224-p224 & $(f)$ web-s112-p224 & $(g)$ concat-s224-p224 & $(h)$ concat-s112-p224 \\
	\end{tabular}
	\caption{Grad-CAM of EfficientNet-B0 trained on different datasets for an image containing pebbles from metis-s224-p224/test}\label{fig:gradCamPebble}
\end{figure}

The most common false positives were due to pictures containing pebbles. Figure \ref{fig:gradCamPebble} depicts the \glspl{gradCam} of the models classifying an image with loose pebbles lying on the ground.
The models were easily confused by loose pebbles, demonstrating that the models learned that pebbles are an important visual cue for honeycombs which corresponds to the definition requiring at least a partially visible pebble. However, the models do not misclassify loose pebbles in general, as some of the examples in the next section illustrate.

In conclusion, \acs{gradCam} confirmed that the metrics-based assessment on which the EfficientNet-B trained \emph{metis-s112-p224} performs best and is the only model which did not develop a bias for the upper left corner.
Surprisingly, the inclusion of the web images did not improve the model's performance. A possible explanation could be that most web images depict excessive honeycombs or that the high preprocessing of these images for web usage is inadequate.

It's also interesting to see that \acs{gradCam} can be used as a tool for instance segmentation, despite only using a classification model as a foundation. 

\subsubsection{Patch Classification}
\begin{figure}[H]
	\centering
	\begin{tabular}{cc}
			\includegraphics[height=4cm]{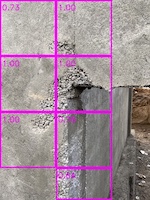}
			&\includegraphics[height=4cm]{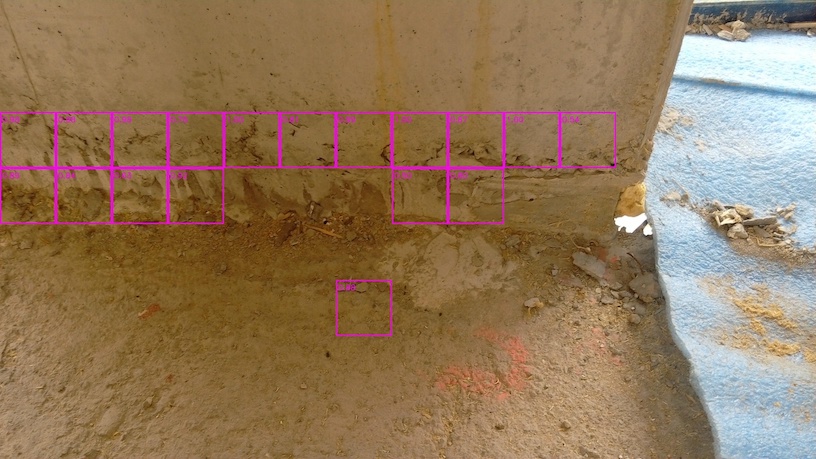}	
			\\
			$(a)$ web example  & $(b)$ metis example\\ 
			\includegraphics[height=4cm]{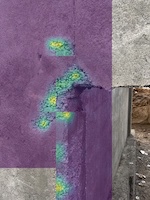}	
			&\includegraphics[height=4cm]{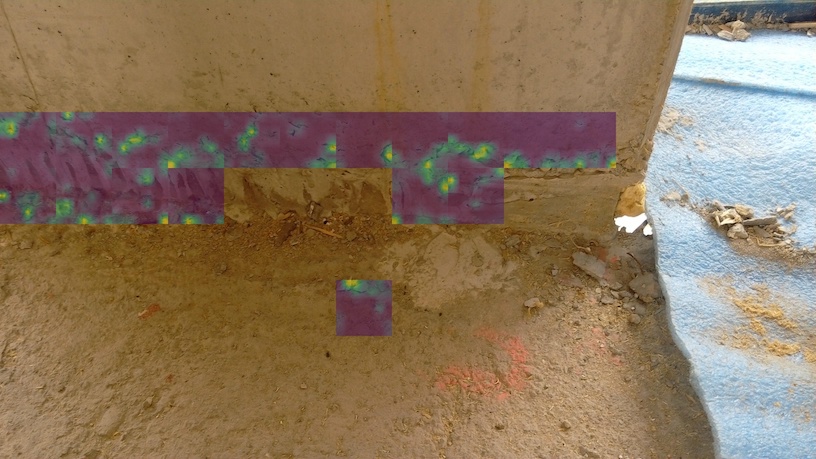}		
			\\
			$(c)$ Grad-CAM overlayed & $(d)$ Grad-CAM overlayed\\
		\end{tabular}
	\caption{$(a)$ and $(b)$ show example images with patch-wise classification. $(c)$ and $(d)$ show the corresponding \glspl{gradCam} activations.}\label{fig:resPatchCam}
\end{figure}

Since the classification model has been trained on lower-resolution patches, we can apply the model patch-wise onto images with larger resolutions to localize defects.
\glspl{gradCam} additionally provides assistance in locating honeycombs for verification. Figure \ref{fig:resPatchCam} demonstrates how \glspl{gradCam} assists in explaining the classification decision and helps localize the honeycomb since the upper left tile in image $(a)$ could be easily misidentified as a false positive. However, the overlayed activation by \glspl{gradCam} shows that the patch was correctly classified due to the small defect in its bottom right corner.
For the two upper images $(a)$ and $(b)$, a magenta border encloses each patch that exceeds the confidence threshold of $0.5$ with the confidence written in the upper left corner of the patch. The two lower images show the corresponding \glspl{gradCam} for the upper images.

If a honeycomb is positioned at the edge of a patch, \acs{gradCam} can provide assistance for human verification as illustrated by $(c)$ and $(d)$ of Figure \ref{fig:resPatchCam}. In the case of patch classification, however, \acs{gradCam} is helpful but mainly serves its initially intended purpose of debugging \glspl{cnn}. That is, recognizing if a model overfits or learns specific undesirable features.
The images scraped from the web seem to lack backgrounds typically seen on construction sites.

While the web dataset only includes images of honeycombs that, for the most part, only contain concrete backgrounds and lack good images of construction sites, the cdc-bhc dataset also includes more differentiated images of concrete. Unfortunately, the models trained on these datasets did not handle typical backgrounds of construction sites sufficiently well, since they are outside the scope of their training data.

\begin{figure}[H]
	\centering
	\begin{tabular}{ccc}
		\includegraphics[height=2.5cm]{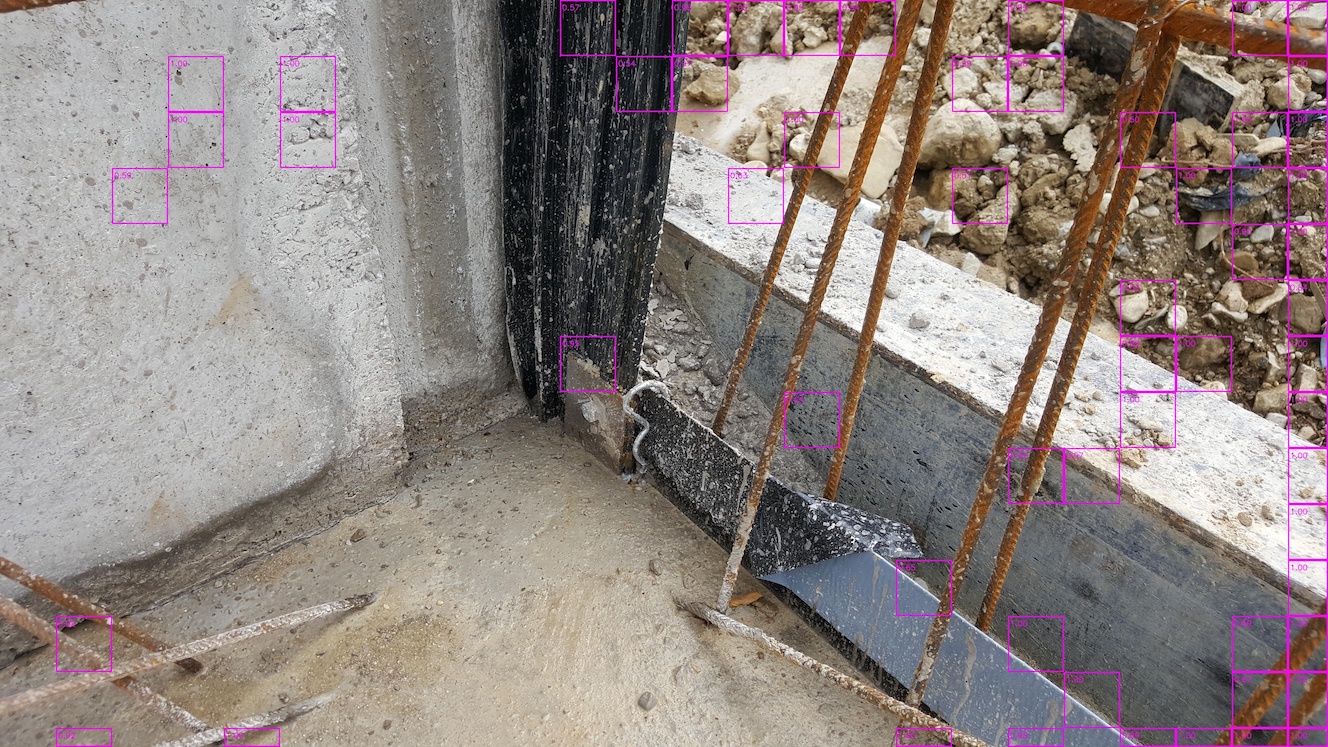}
		&\includegraphics[height=2.5cm]{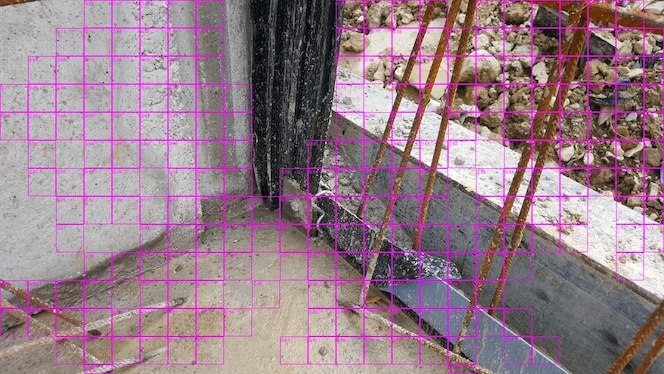}	
		&\includegraphics[height=2.5cm]{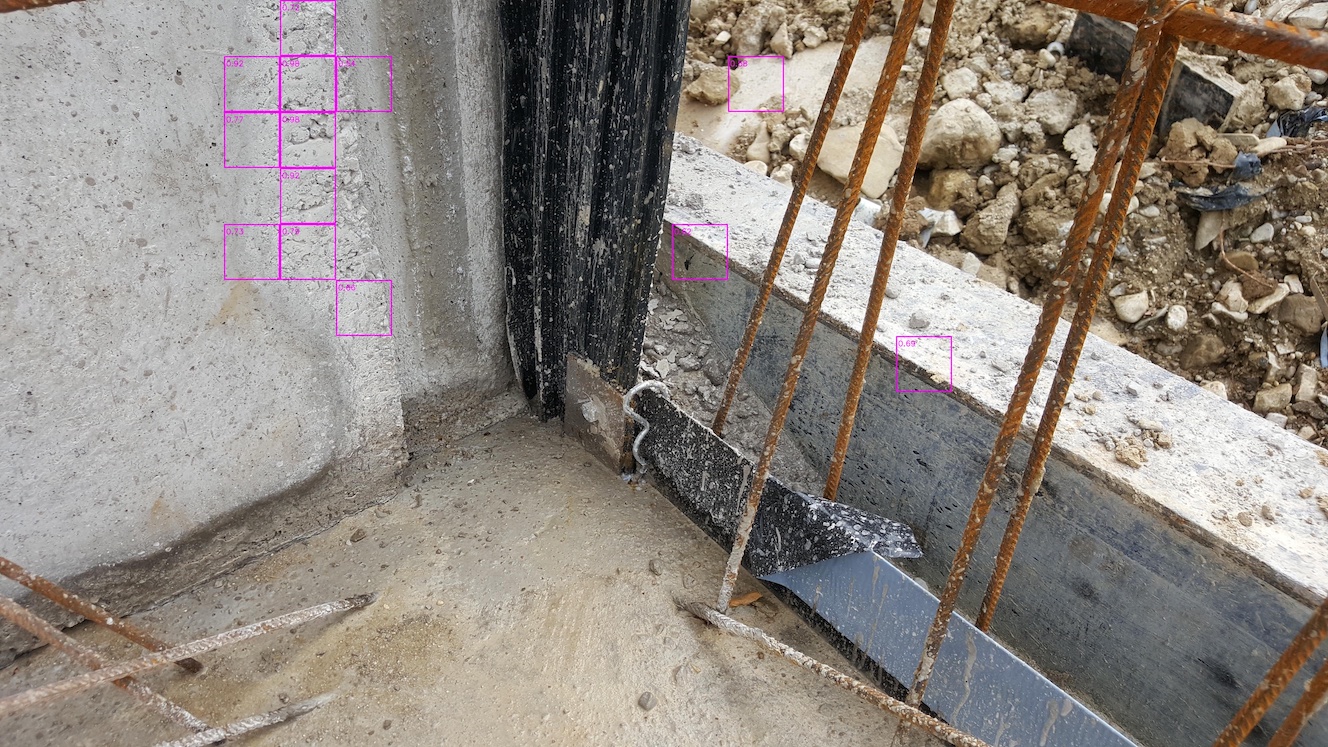}		
		\\
        $(a)$ cdc-bhc & $(b)$ web-s224-p224 & $(c)$ metis-s122-p224
	\end{tabular}
	\caption{False positives for non-concrete patches by our finetuned EfficientNet-B0 trained on different training sets}\label{fig:gradCamExample4}
\end{figure}

Figure \ref{fig:gradCamExample4} illustrates the higher false positive rates, particularly for areas not depicting concrete surfaces. The model trained on the \emph{web} data performed particularly poorly in this case, even worse than the \emph{cdc-bhc} model, although the performance metrics are higher for the web model. However, this is to be expected since the \emph{cdc-bhc} dataset contains a wider variety of negative examples.
In conclusion, adding more images depicting construction sites may further decrease the false positive rate.

\begin{figure}[H]
	\centering
	\begin{tabular}{ccc}
		\includegraphics[height=2.5cm]{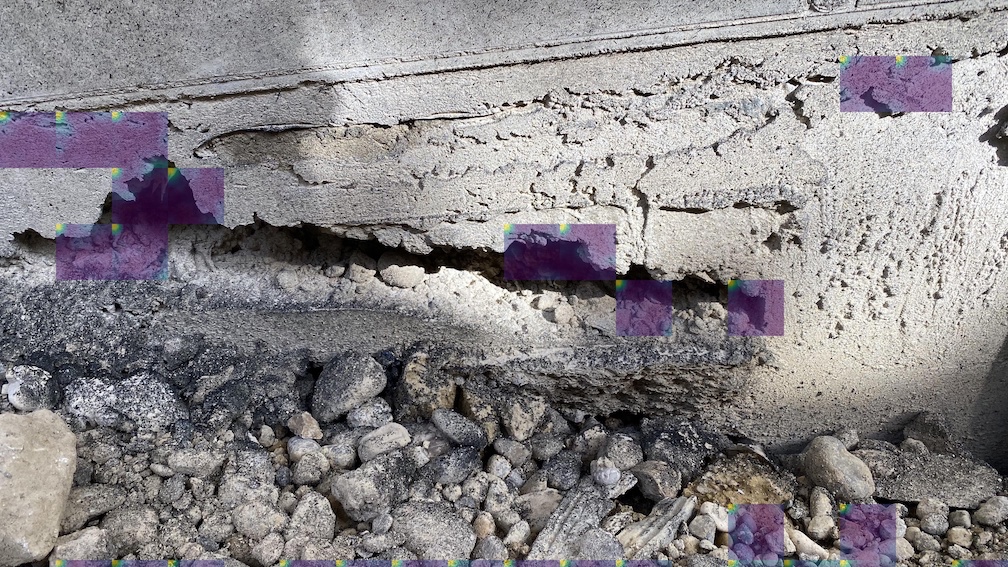}
		&\includegraphics[height=2.5cm]{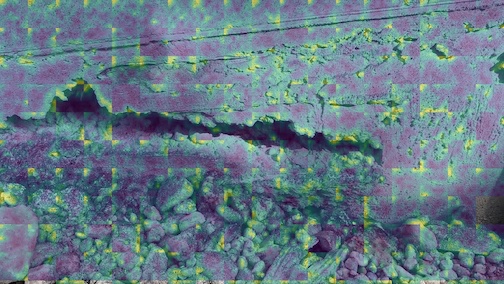}	
		&\includegraphics[height=2.5cm]{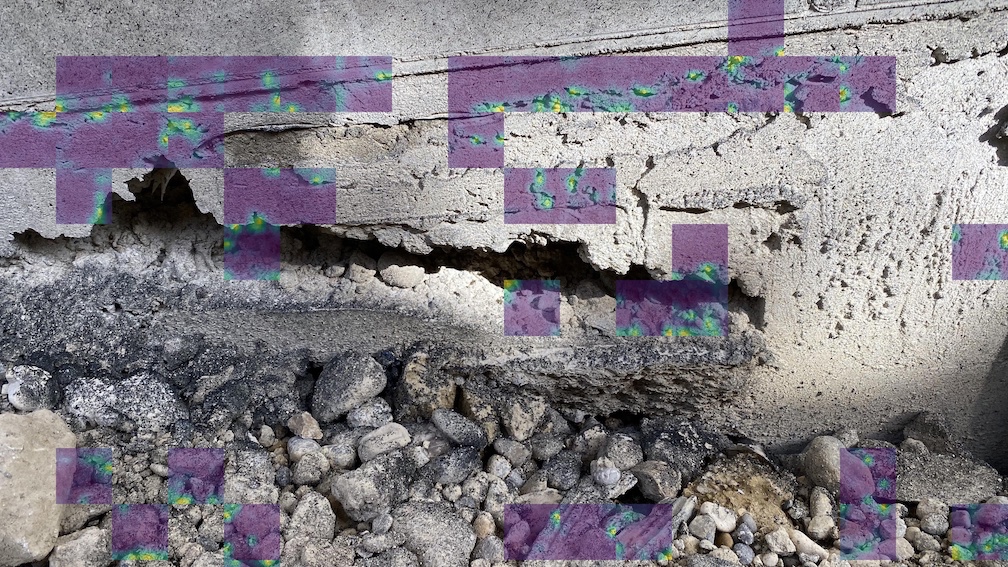}		
		\\
        $(a)$ cdc-bhc & $(b)$ web-s224-p224 & $(c)$ metis-s122-p224
	\end{tabular}
	\caption{Untypical scaling of a honeycomb by our finetuned EfficientNet-B0 trained on different training sets}\label{fig:resCamLarge}
\end{figure}

Since none of the applied augmentations targeted differences in the input's scaling, the models naturally could not learn honeycomb structures that are too large, as shown by Figure \ref{fig:resCamLarge} with an atypically close photo of a honeycomb. However, the classification model naturally failed to recognize most patches correctly as honeycombs since neither the classification model nor the training data addressed significant differences in size.
This weakness in handling unexpected scales could be addressed by generating the patches from the image pyramids \cite{adelson1984pyramid}. For instance, Xiao \acs{etal}\ \cite{xiao2020surface} added image pyramids to \gls{mrcnn}, which improved the performance slightly. However, Girshick \cite{girshick2015fast} argued that the improvements gained by using image pyramids are not significant enough to justify the increase in computation time for \emph{Fast R-CNN}, and Ren \acs{etal}\ \cite{ren2015faster} argued the same for \emph{Faster R-CNN}. Since Girshick \cite{girshick2015fast} also demonstrated that the model \emph{Fast R-CNN} learns scale invariance from the training data, increasing the scale variance by adding patches of image pyramids might improve the classification model.

\subsection{Instance segmentation}
The \gls{mrcnn} models learned to detect honeycombs to a certain degree.
 
\begin{figure}[H]
	\centering
		\includegraphics[width=0.45\linewidth]{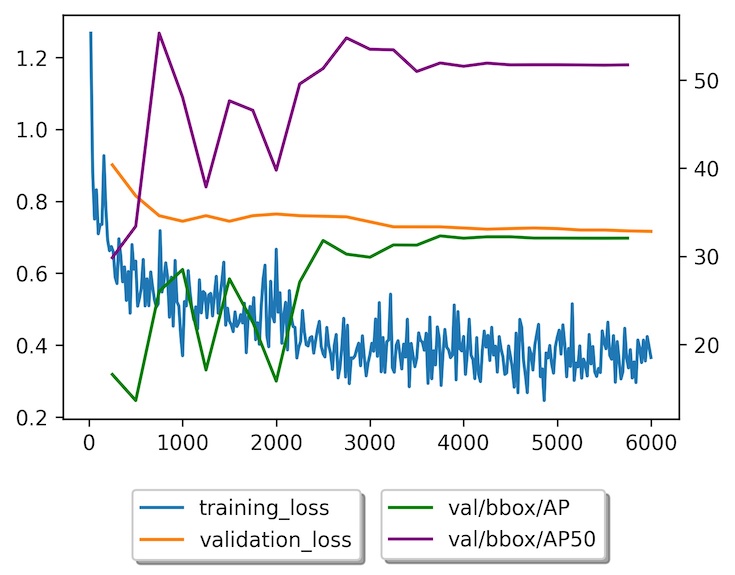}
		\includegraphics[width=0.45\linewidth]{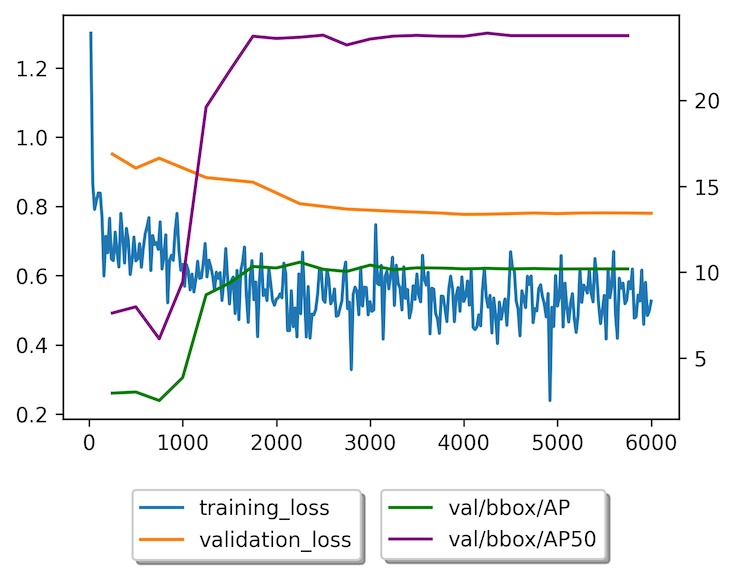}
		 \includegraphics[width=0.45\linewidth]{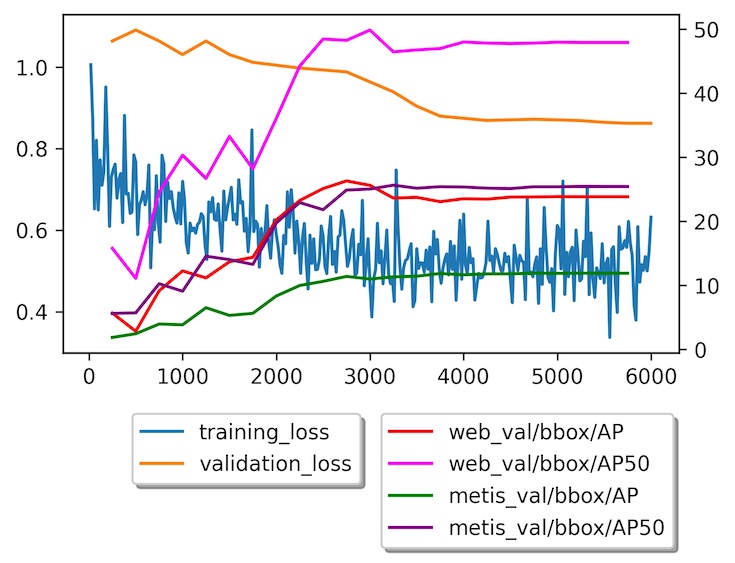}
		\\
		$(a)$  $(b)$  $(c)$ \\
	\caption{Mask R-CNN training}\label{fig:mrcnnTraining}
\end{figure}

Figure \ref{fig:mrcnnTraining} depicts the training and validation losses as well as the bounding box \glspl{ap} on the validation set. 

Both models trained on either dataset did not improve significantly past iteration 3000, while the model trained on both datasets converged at around 4000 iterations.
The validation \gls{ap} jumped significantly for the \emph{web} validation set compared to the \emph{metis} validation set caused by the small size of the \emph{web} dataset compared to the \emph{metis} dataset.

Table \ref{tab:resBbox} displays the metrics of the \gls{mrcnn} model trained on the \gls{hicis} \emph{web} and \emph{metis} datasets, as well as the combination of both achieved on the validations sets. 
\begin{table}[H]
	\centering
	\begin{tabular}{!{\extracolsep{4pt}}lcccccc}
		& \multicolumn{3}{c}{\textbf{Web}} & \multicolumn{3}{c}{\textbf{Metis}} \\
		\cline{2-4}\cline{5-7}
		\textbf{metric} & \textbf{W} & \textbf{M} & \textbf{W+M} & \textbf{W} & \textbf{M} & \textbf{W+M} \\
		
		\toprule
		$AP_{IoU \geq 50}$ 					& \textbf{37.4} & 16.4 & 25.6 & 8.9 & 12.3 & \textbf{12.4} \\
		$AP_{IoU \geq 0.5 : 0.95 : 0.05}$ 	& \textbf{22.2} & 7.9 & 17.4 & 3.1 & \textbf{6.0} & 5.7 \\
		$AR_{IoU \geq 0.5 : 0.95 : 0.05}$ 	& \textbf{28.2} & 9.6 & 18.6 & 7.6 & 8.1 & \textbf{8.8} \\
		
	\end{tabular}
	\caption{Metrics for bounding boxes}\label{tab:resBbox}
\end{table}

Table \ref{tab:resMask} presents the corresponding metrics for the instance segmentation masks.
The \emph{web} model achieved lower performance on the \emph{metis} dataset than the \emph{metis} model, although achieving a slightly higher \gls{ar}. However, the inclusion of the \emph{web} data improved the model's segmentation masks slightly.

\begin{table}[H]
	\centering
	\begin{tabular}{!{\extracolsep{4pt}}lcccccc}
		& \multicolumn{3}{c}{\textbf{Web}} & \multicolumn{3}{c}{\textbf{Metis}} \\
		\cline{2-4}\cline{5-7}
		\textbf{metric} & \textbf{W} & \textbf{M} & \textbf{W+M} & \textbf{W} & \textbf{M} & \textbf{W+M} \\
		
		\toprule
		$AP_{IoU \geq 50}$ 					& \textbf{33.0} & 14.5 & 23.2 & 7.3 & 11.7 & \textbf{11.9} \\
		$AP_{IoU \geq 0.5:0.95:0.05}$ 	& \textbf{17.2} & 6.7 & 15.9 & 2.5 & 4.1 &\textbf{4.4} \\
		$AR_{IoU \geq 0.5:0.95:0.05}$ 	& \textbf{23.7} & 8.6 & 17.0 & 6.1 & 6.0 & \textbf{7.0} \\
		
	\end{tabular}
	\caption{Metrics for segmentation masks}\label{tab:resMask}
\end{table}

\begin{figure}[H]
	\centering
	\begin{tabular} {cc}
		\includegraphics[height=6cm]{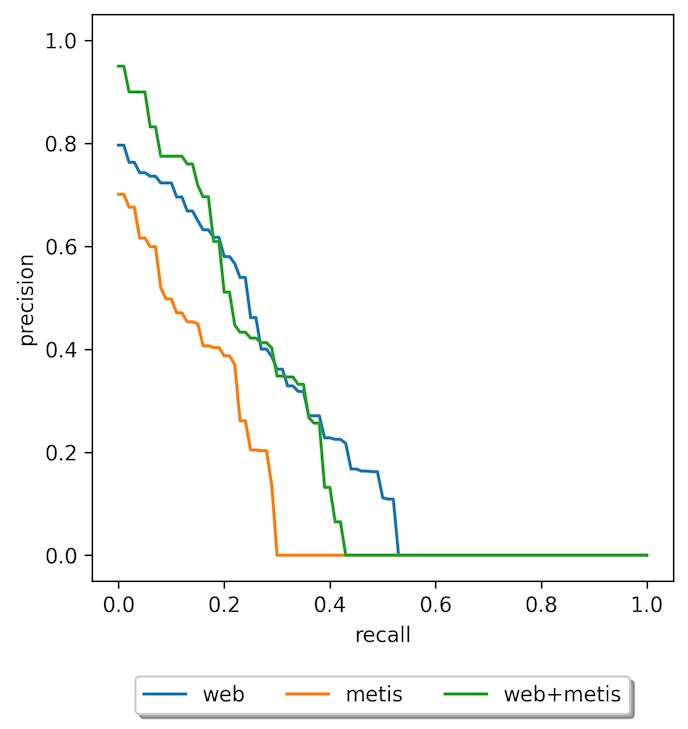}
		&\includegraphics[height=6cm]{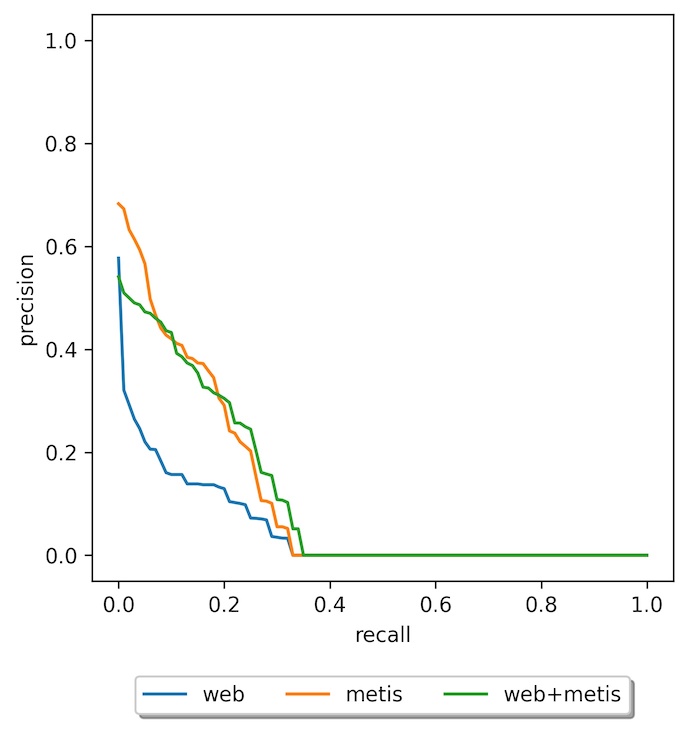}
		\\
		$(a)$ \emph{web} test set & $(b)$ \emph{metis} test set \\
	\end{tabular}
	\caption{Precision-recall curves}\label{fig:resPrCurveTest}
\end{figure}

All models achieved higher scores on the \emph{web} test set independent of the training set combination.
The inclusion of the \emph{metis} training data decreased the performance on the \emph{web} test set. The significantly more extensive size of the \emph{metis} dataset compared to the \emph{web} dataset led to a higher emphasis on the realistic images, causing the model trained on both datasets to perform worse on the \emph{web} dataset. Therefore, the decreased performance affirms the assumption that the images scraped from the web represent the most easily recognizable honeycombs.
The inclusion of the \emph{web} training set improved the model slightly on the \emph{metis} dataset, achieving the best values out of the three models in nearly all metrics.
Furthermore, the model trained on both datasets outperformed the model trained on only the \emph{metis} dataset for recall values over 20\% as illustrated by Figure \ref{fig:resPrCurveTest}.

In conclusion, the two datasets differ significantly, with the web images representing a limited selection of honeycombs.

Figure \ref{fig:resPrCurveTest} $(a)$ and $(b)$ depict the precision-recall curves of all three models using an \gls{iou} threshold of $0.5$, respectively, on the \emph{web} test set and the \emph{metis} test set. The two precision-recall curves illustrate the cause of low $AP$ and $AR$. Since the precision or recall is zero at many thresholds, the averages of those are significantly lowered, caused by the model failing to achieve precision and recall values above a certain point regardless of the confidence threshold. 

In summary, metrics for bounding boxes and segmentation masks showed a performance improvement when combining both datasets, indicating that increasing the variance in the dataset by increasing its size is likely to lead to further increases in performance.

\subsubsection{Evaluation for practical application}
Since considering only the average precision and recall metrics might lead to underestimating the model's performance, the precision and recall at a specific confidence threshold were calculated. Therefore, the specification of a suitable value for the confidence threshold was required.

Table \ref{tab:valConfPr} shows the metrics of all models on both validation sets using an \gls{iou} threshold of 50\% at three different confidence thresholds for the bounding box detections. For instance, with a confidence threshold of 0.7, the model achieves a precision of 72.5\%, which is vastly higher than expected when just considering the \gls{ap}.

\begin{table}[H]
	\centering
	\scriptsize
	\begin{tabular}{!{\extracolsep{4pt}}rccccccccc}
		\multicolumn{2}{c}{} & \multicolumn{4}{c}{\textbf{Web}} & \multicolumn{4}{c}{\textbf{metis}} \\
		\cline{3-6} \cline{7-10}
		\addlinespace[1ex] \textbf{model} & \textbf{confidence}  & \textbf{precision} &  \textbf{recall} & \textbf{f1-score} & \textbf{support} & \textbf{precision} &  \textbf{recall} & \textbf{f1-score} & \textbf{support} \\
		\toprule
		\addlinespace[1ex] \multicolumn{1}{c|}{\multirow{3}{*}{\textbf{web}}} 
		 & 0.3 & 0.3469 & 0.4722 & 0.4000 & 36 & 0.1386 & 0.2121 & 0.1677 & 198 \\
		 \multicolumn{1}{c|}{}& 0.5 & 0.4706 & 0.4444 & 0.4571 & 36& 0.1954 & 0.1799 & 0.1873 & 189 \\
		 \multicolumn{1}{c|}{}& 0.7 & 0.6087 & 0.3889 & 0.4746 & 36& 0.2526 & 0.1379 & 0.1784 & 174 \\
		
		\addlinespace[1ex] \multicolumn{1}{c|}{\multirow{3}{*}{\textbf{metis}}} 
		& 0.3 & 0.5333 & 0.4444 & 0.4848 & 36 & 0.2060 & 0.3179 & 0.2500 & 195 \\
		\multicolumn{1}{c|}{}& 0.5 & 0.7857 & 0.3056 & 0.4400 & 36 &  0.3077 & 0.2051 & 0.2462 & 195 \\
		\multicolumn{1}{c|}{}& 0.7 & 0.7500 & 0.3000 & 0.4286 & 30 & 0.4694 & 0.1447 & 0.2212 & 159 \\

		\addlinespace[1ex] \multicolumn{1}{c|}{\multirow{3}{*}{\textbf{W+M}}} 
		& 0.3 &0.6429 & 0.5000 & 0.5625 & 36 & 0.2576 & 0.3434 & 0.2944 & 198 \\
		\multicolumn{1}{c|}{}& 0.5 &  0.8500 & 0.4722 & 0.6071 & 36 & 0.4000 & 0.2383 & 0.2987 & 193 \\
		\multicolumn{1}{c|}{}& 0.7 & 0.9286 & 0.3611 & 0.5200 & 36 & .5283 & 0.1637 & 0.2500 & 171 \\
	\end{tabular}
	\caption{Differing confidence thresholds at an $IoU=0.5$ on the validation sets}\label{tab:valConfPr}
\end{table}

Table \ref{tab:testConfPr} shows the corresponding metrics achieved on the test sets compared to the validation set as shown by Table \ref{tab:valConfPr}, since the threshold must be determined on the validation set. Otherwise, there would not be sufficient evidence that the determined threshold also generalizes adequately.
\begin{table}[H]
	\centering
	\scriptsize
	\begin{tabular}{!{\extracolsep{4pt}}cccccccccc}
		\multicolumn{2}{c}{} & \multicolumn{4}{c}{\textbf{web}} & \multicolumn{4}{c}{\textbf{metis}} \\
		\cline{3-6} \cline{7-10}
		\addlinespace[1ex] \textbf{model} & \textbf{confidence}  & \textbf{precision} &  \textbf{recall} & \textbf{f1-score} & \textbf{support} & \textbf{precision} &  \textbf{recall} & \textbf{f1-score} & \textbf{support} \\
		\toprule
		\addlinespace[1ex] \multicolumn{1}{c|}{\multirow{3}{*}{\textbf{web}}} 
		& 0.3 &  0.3396 & 0.6316 & 0.4417 & 57 & 0.2222 & 0.3143 & 0.2604 & 210 \\
		\multicolumn{1}{c|}{}& 0.5 & 0.4225 & 0.5263 & 0.4687 & 57 & 0.3161 & 0.3216 & 0.3188 & 171\\
		\multicolumn{1}{c|}{}& 0.7 & 0.5682 & 0.4386 & 0.4950 & 57 & 0.3711 & 0.2323 & 0.2857 & 155 \\
		
		\addlinespace[1ex] \multicolumn{1}{c|}{\multirow{3}{*}{\textbf{metis}}} 
		& 0.3 & 0.4167 & 0.4386 & 0.4274 & 57 & 0.3162 & 0.4095 & 0.3568 & 210 \\
		\multicolumn{1}{c|}{}& 0.5 & 0.6800 & 0.2982 & 0.4146 & 57 &  0.4911 & 0.3293 & 0.3943 & 167 \\
		\multicolumn{1}{c|}{}& 0.7 & 0.8333 & 0.1923 & 0.3125 & 52 & 0.7250 & 0.2266 & 0.3452 & 128 \\

		\addlinespace[1ex] \multicolumn{1}{c|}{\multirow{3}{*}{\textbf{W+M}}} 
		& 0.3 &0.4304 & 0.5965 & 0.5000 & 57 & 0.3116 & 0.4115 & 0.3546 & 209 \\
		\multicolumn{1}{c|}{}& 0.5 &  0.6486 & 0.4211 & 0.5106 & 57 & 0.4786 & 0.3415 & 0.3986 & 164 \\
		\multicolumn{1}{c|}{}& 0.7 &0.8333 & 0.2632 & 0.4000 & 57 & 0.6531 & 0.2462 & 0.3575 & 130 \\
	\end{tabular}
	\caption{Differing confidence thresholds at an $IoU=0.5$ on the test sets}\label{tab:testConfPr}
\end{table}
The confidence threshold of 0.5 was confirmed by the $F_1$ score on the \emph{metis} test set. Each model reaches its highest $F_1$ score on the \emph{web} test set at different confidence thresholds. However, the support is less than for the \emph{metis} set, and the difference in $F_1$ scores was minor, making it likely that the \emph{web} test set was not large enough. Overall the metrics confirm the selection of a confidence threshold of 0.5 based on the validation set.

\begin{figure}[H]
	\centering
	\begin{tabular}{ccc}
		\includegraphics[height=4cm]{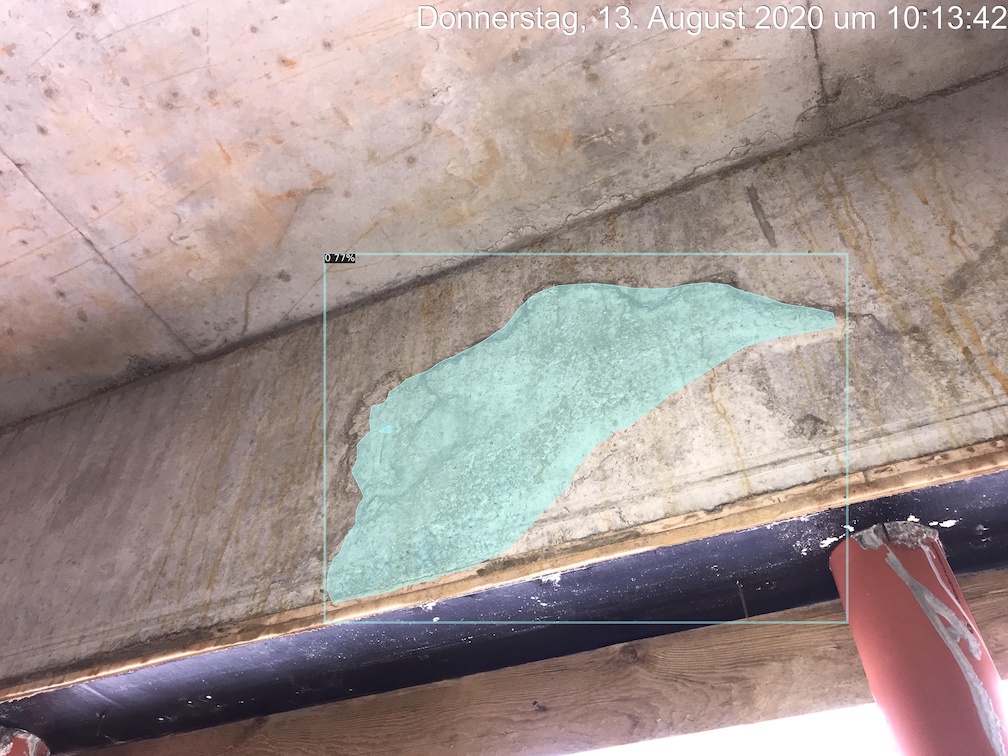}
		& \includegraphics[height=4cm]{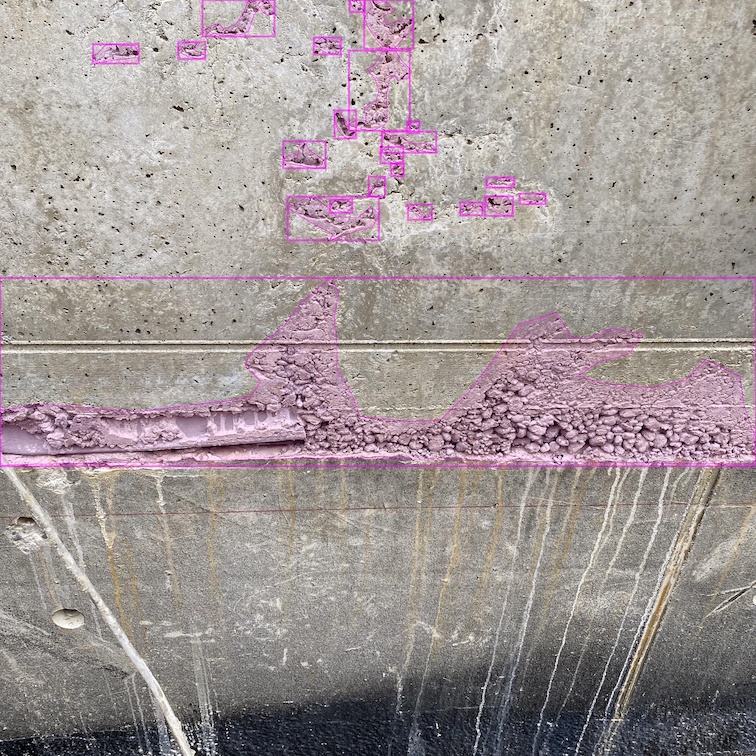}
		& \includegraphics[height=4cm]{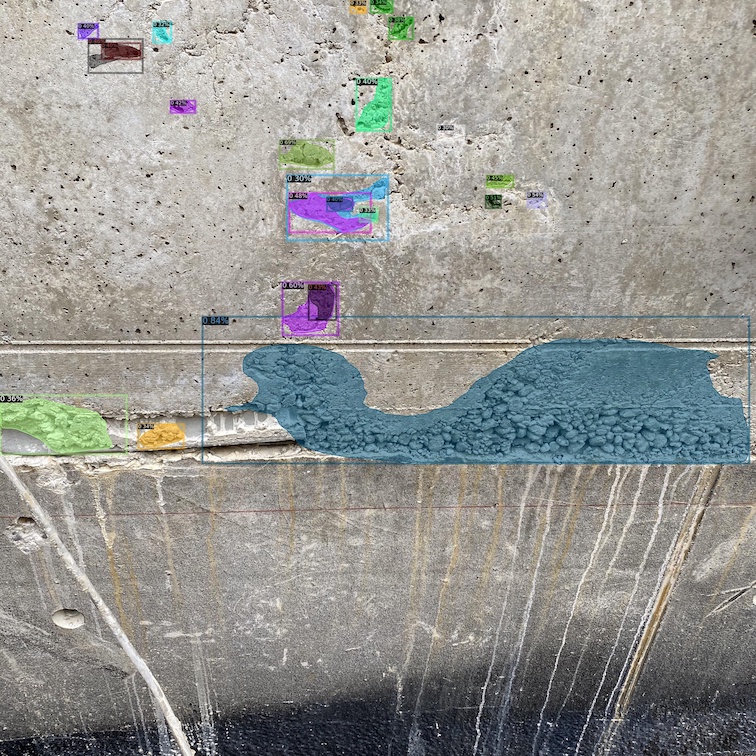}
		\\
		$(a)$ near perfect segmentation mask & $(b)$ ground truth & $(c)$ predictions\\
	\end{tabular}
	\caption{Near perfect segmentation mask and differing division of honeycombs}\label{fig:resQualityHigh}
\end{figure}

The high quality of some of the model's detection rates is illustrated by image $(a)$ of Figure \ref{fig:resQualityHigh}. The model predicted a near-perfect segmentation mask for the honeycomb in the image. However, this is not representative of the detection quality in general.

As illustrated by images $(b)$ and $(d)$ in Figure \ref{fig:resQualityHigh}, the model's predictions dividing honeycombs into single or multiple instances reflect the challenges of defining instances of honeycombs as discussed in section \ref{sec:hicis}. Furthermore, the difference in dividing honeycombs results in fewer true positives since the split honeycomb instances do not reach an \gls{iou} of $0.5$. One could address this issue by changing the problem type to classification and segmentation. Although an experiment classifying large honeycombs failed, as mentioned in section \ref{sec:hicc}, the challenge of unclear instance divisions may justify further research in this direction. Nevertheless, this problem does not affect verification by inspectors and, therefore, might be addressed by increasing the training data.

\begin{figure}[H]
	\begin{tabular}{ccc}
		\includegraphics[width=0.3\textwidth]{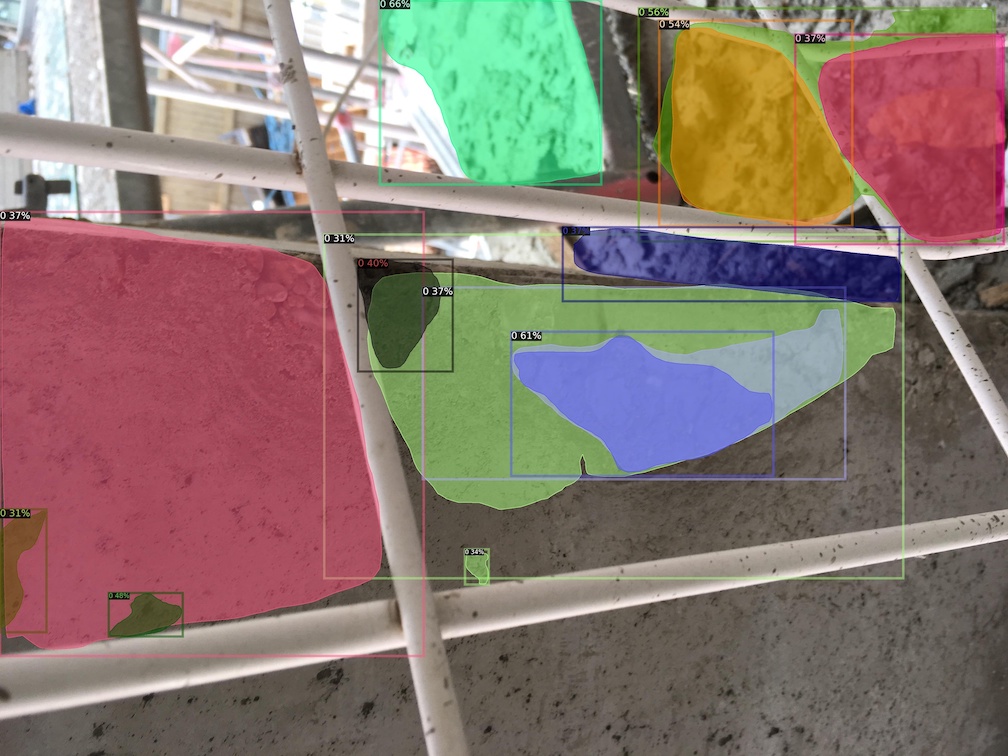}
		\includegraphics[width=0.3\textwidth]{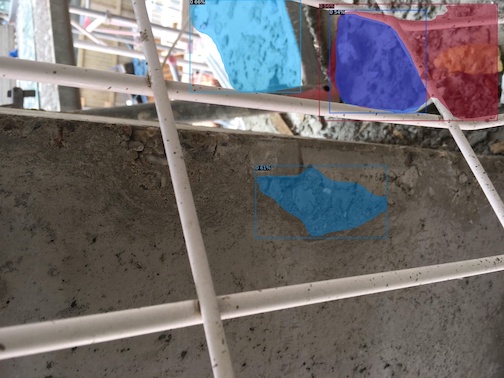}
		\includegraphics[width=0.3\textwidth]{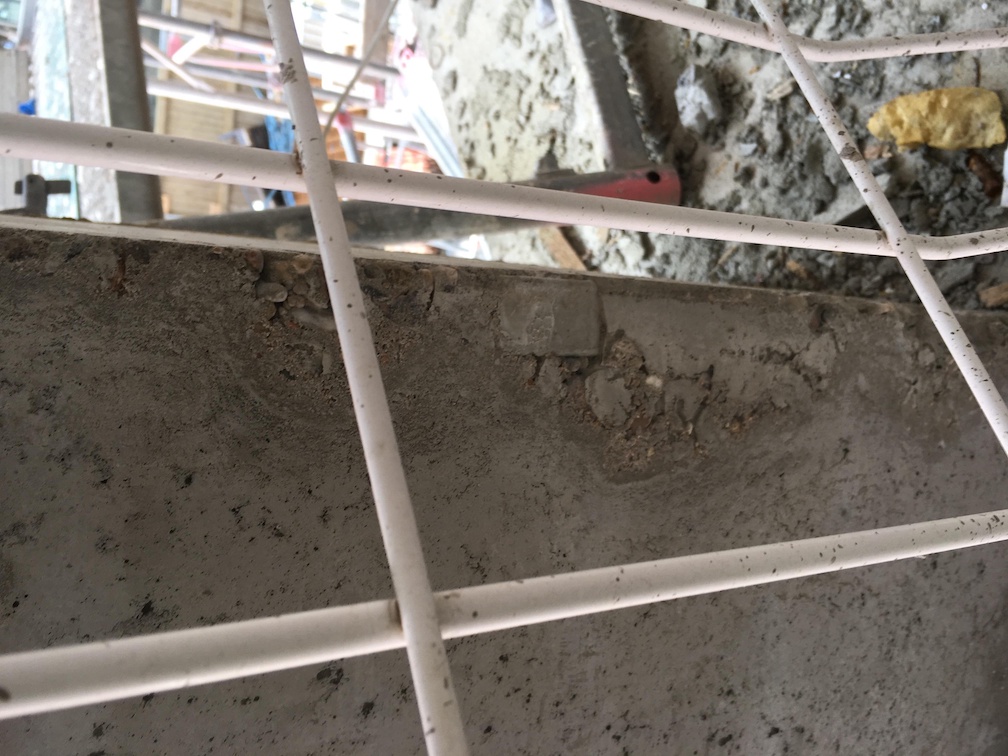}
		&& \\
	\end{tabular}
	\caption{Lose pebbles detected as honeycombs}\label{fig:fpPebbles}
\end{figure}

While the detections at a confidence threshold of $0.3$ are mostly correct, lowering the confidence threshold increased the number of false positives when encountering loose pebbles on the ground, as shown in Figure \ref{fig:fpPebbles}.

In conclusion, the model performed reasonably well and shows promise for object detection to target honeycombs and possible defects with a vague outline. However, the differentiation between pebbles and honeycombs needs more research. Two possible approaches are the segmentation of concrete surfaces or adding more images of loose pebbles. Bunrit \acs{etal}\ \cite{bunrit2019evaluating} demonstrated that materials used in construction, such as concrete, could be classified. 
Furthermore, the division of honeycombs into the correct instances may be a challenge, although likely to be solved by using an adequately large dataset.

\subsection{Instance Segmentation vs. Classifying Patches}

The previous two sections explored and discussed the results of the object detection and classification approaches. While object detection and classification metrics cannot be compared directly, they may still indicate differences.

Comparing the classification model trained on \emph{\gls{hicc}-metis-s224-p224} to the \gls{mrcnn} trained on the \gls{hicis} dataset, the classification model can achieve higher precision and recall. However, considering the harshness of the \gls{iou}, the difference is not large enough to dismiss the performance of the \gls{mrcnn} model. Another consideration is that users may perceive bounding boxes with segmentation masks as more natural compared to patch classification with \acs{gradCam} as \acs{eg} bounding circles are commonly used to indicate an area of importance in an image.

The labeling of images for instance segmentation requires vastly more effort than classification. Instance masks address a much more difficult problem formulation than class labels: The location of an instance, which pixels belong to an instance, and the number of instances in the image. While the classification of a large image may not adequately address the user's requirement to locate the defect for faster verification, slicing an image into classifying patches may solve this problem, especially considering the effort of data acquisition.

\subsubsection{Qualitative Analysis on Realistic Images}
The validity of the quantitative comparison between instance segmentation and patch-based classification is limited because both address the problem in fundamentally different ways.
Although comparing object detection and patch-based classification using the \gls{iou} is possible \cite{veeranampalayam2020comparison}, this approach reduces the problem type to segmentation, which is not the initial problem type for either classification.
Since the quantitative comparison between the instance segmentation and patch-based classification approaches was not sufficient, an expert in defect documentation appraised the predictions manually, enabled by the small size of the test set. 

The qualitative analysis compared the detections and patch classifications of the best models side by side, \acs{ie} the \gls{mrcnn} trained on both datasets and the EfficientNet-B0 trained on only the \gls{hicc} \emph{metis-s112-p224} datasets respectively. For each test image, the following criteria were applied. First, it was checked whether the crucial honeycomb was detected, meaning the largest in most cases. Second, it was checked if other significant honeycombs were detected. Third, it was checked how many false positives existed and if the number exceeded the true positive detection count. Then an assessment of unsatisfactory, sufficient, or satisfactory was given. \emph{Satisfactory} describes an image with honeycomb detections that a potential user could confirm by a glance requiring at least one correct detection. \emph{Sufficient} expresses that an image with detections could not be verified easily, either due to too many false positives or failure to detect the crucial honeycomb. Finally, \emph{Unsatisfactory} describes the detection for an image that would have been unusable either due to it failing to detect any honeycomb or failing to detect any honeycomb correctly.

Since a direct comparison by metrics is not possible between classification and object detection, a qualitative analysis was necessary, although its results are limited by subjective bias.

Table \ref{tab:resQualScores} summarizes the assessment. While the instance segmentation approach produced more satisfactory detections, classifying patches had fewer unsatisfactory results.
\begin{table}[H]
	\centering
	\begin{tabular}{rccc}
		\textbf{model}		& \textbf{unsatisfactory} & \textbf{sufficient} & \textbf{satisfactory} \\
		\toprule
		EfficientNet-B0 trained on \emph{metis-s112-p224}	& 4 & 22 & 12 \\
		\gls{mrcnn} trained on both datasets          	& 8 & 12 & 18 \\
	\end{tabular}
	\caption{Assessment of detecting honeycombs in 38 realistic test images}\label{tab:resQualScores}
\end{table}

When directly comparing the detections, 16 detections were of similar quality. The object detection approach achieved better results in 13 cases but performed worse for eight images. 
The comparison confirmed that the instance segmentation masks are more convenient for users to validate.

\begin{figure}
	\centering
	\begin{tabular}{cc}
		\includegraphics*[width=0.3\linewidth]{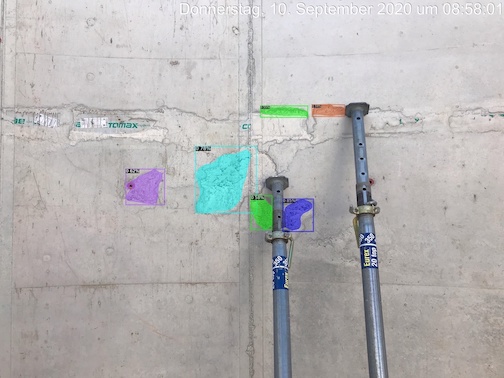}
		& \includegraphics*[width=0.3\linewidth]{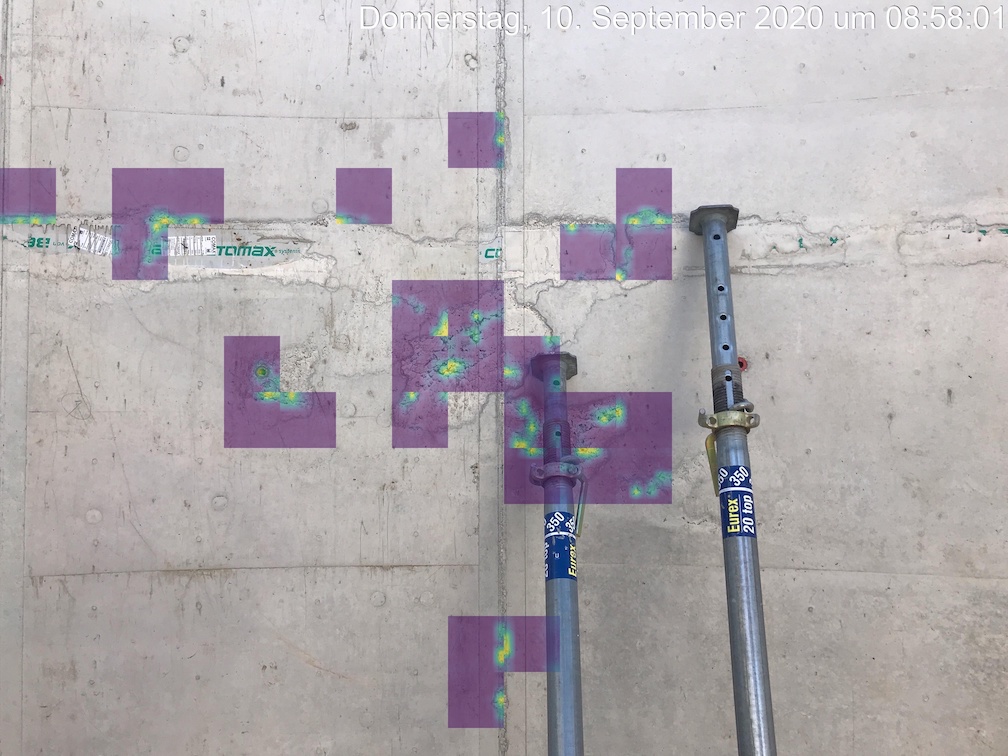}
		\\
		$(a)$ instance segmentation & $(b)$ classifying patches \\
	\end{tabular}

	\caption{Intuitiveness of instance segmentation compared to patch classification}\label{fig:easyConfirm}
\end{figure}

Figure \ref{fig:easyConfirm} demonstrates the advantages of instance segmentations since the segmentations are more distinctive, especially around the steel pipes.
However, this might be addressed by classifying sliding patches, \acs{eg} a slide of 0.5 of the patch size. When decreasing the slide to a pixel, each pixel is classified, which is essentially segmentation. However, this segmentation would be missing the instance identifications of the \gls{mrcnn} model.
 
\begin{figure}
	\centering
	\begin{tabular}{cc}
		\includegraphics*[width=0.3\linewidth]{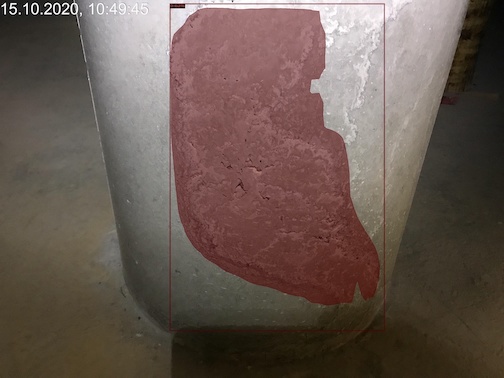}
		& \includegraphics*[width=0.3\linewidth]{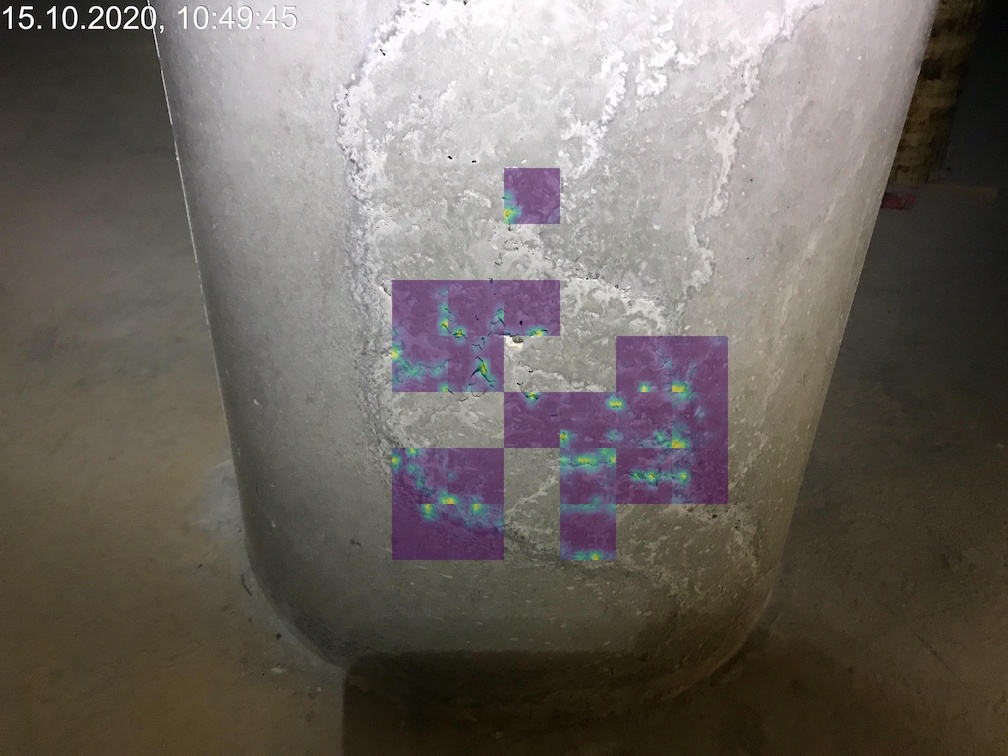} \\
		$(a)$ & $(b)$ \\
		\includegraphics*[width=0.3\linewidth]{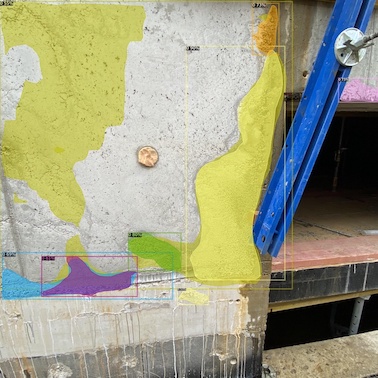}
		& \includegraphics*[width=0.3\linewidth]{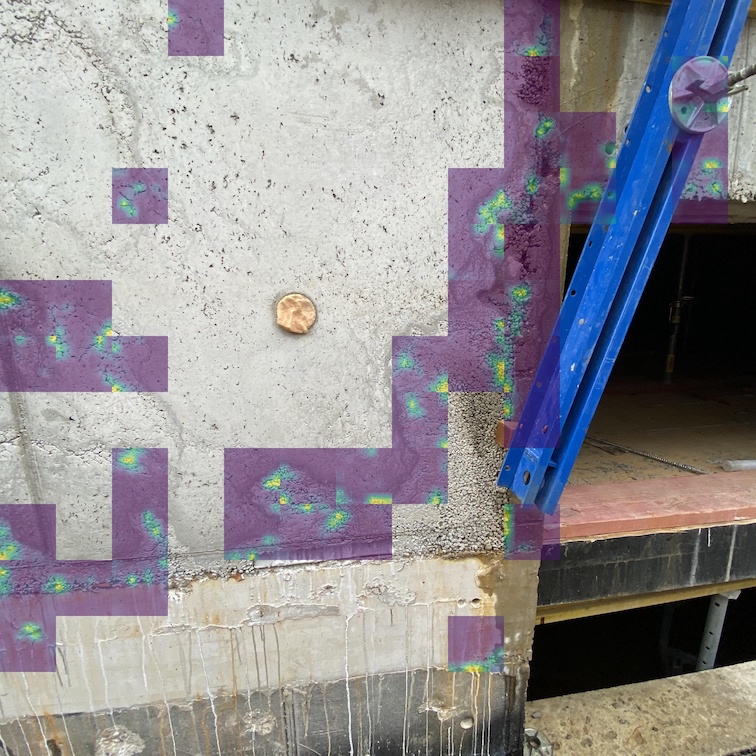} \\
		$(c)$ & $(d)$ \\
	\end{tabular}
	\caption{Continuity of detections}\label{fig:detectContinuouity}
\end{figure}

Classifying by patches tended to create discontinuous honeycombs, while \gls{mrcnn} produced more continuous detections as illustrated by Figure \ref{fig:detectContinuouity}.

In conclusion, both instance segmentation and classification with \acs{gradCam} are valid approaches for further research with larger datasets. Since the labeling of honeycombs for instance segmentation is significantly more challenging than classification, the latter could be superior for future research, especially considering the expert knowledge required and the ambiguities dividing honeycombs.  
However, since instance segmentations are more natural to potential users and an initial \gls{mrcnn} model could be trained, an active learning approach could reduce labeling effort as demonstrated by Feng \acs{etal}\ \cite{feng2017deep}.
Therefore, future research should look into the problem using either approach depending on the integration of an active learning framework into a defect documentation system.

%% file: Chapter_Conclusion.tex
\section{Conclusion}
\label{chap:conclusion}

\subsection{EfficientNet-B0}
An EfficientNet-B0 was trained on the \gls{cdc} dataset \cite{hung2019surface} without adding any additional augmentations.
The model achieved with $96.95\%$ accuracy, $97.32\%$ precision, and $96.76\%$ recall state-of-the-art on the \gls{cdc} dataset.
The successful application of EfficientNet-B0 demonstrated that models' improvements on ImageNet can also improve performance for transfer-learning. Furthermore, it confirmed its adequacy for training a binary honeycomb classifier.

\subsection{HiCIS and HiCC datasets}
The \gls{hicc} and \gls{hicis} datasets are published on GitHub for use in research: \url{https://github.com/jdkuhnke/HiC}. \Gls{hicc} contains binary classification datasets for honeycombs in concrete. \Gls{hicis} contains datasets for detecting honeycombs with bounding boxes and instance segmentation masks labeled in the \emph{MS COCO} format. These datasets provide a basis for further research into honeycomb detection. The raw images are also included.

While the instance segmentation masks were labeled in good faith, smaller honeycombs may not get labeled accurately enough for the 224 x 224 pixels patches in large images with fractured honeycombs. As a result, it is likely that the \gls{hicc} dataset likely contains some incorrect class labels.

The \gls{hicis} dataset could be extended by adding more challenging honeycombs from the raw images supplied in the repository. 
The labeling process could be eased by training an initial model and using an active learning approach. The model would continuously train on data and create predictions for new unlabeled data, then feed back into the model for additional training. Feng \acs{etal}\ \cite{feng2017deep} demonstrated that active learning could assist with labeling defect images.

\subsection{Differences in datasets obtained from real scenarios and scraped from the web}
Differences between honeycomb datasets scraped from the web and images obtained from real scenarios were explored.
Both experiments, \acs{ie} classification and detection, suggest that images scraped from the web and those obtained from the field differ substantially.

In the case of classifying patches, the EfficientNet-B0 trained on the \gls{hicc} \emph{web-s224-p224} dataset achieved significantly lower scores on the \emph{metis-s224-p224} test set, \acs{ie} 41.3\% \gls{ap} and 42.5\% \gls{ar}, compared to the model trained on \emph{metis-s112-p224} achieving 68.0\% \gls{ap} and 67.2\% \gls{ar}. Furthermore, the model trained on \emph{metis-s112-p224} achieved an 97.3\% \gls{ap} and 96.3\%  \gls{ar} on the \emph{web-s224-p224} test set close to its scores of the model trained on \emph{web-s224-p224} , achieving 98.7\% \gls{ap} and 97.8\% \gls{ar}.

For the instance segmentation, the \gls{mrcnn} trained on any combination of \gls{hicis} training datasets performed better on the \emph{web} test set than the \emph{metis} test set. The different performance on both datasets indicates a difference in the distribution of the two datasets.

In conclusion, although the dataset scraped from the web does not fully represent the complete variance of honeycombs, it still represents a limited selection. However, the inclusion of web images did not necessarily improve the accuracy of the model on realistic images.

\subsection{Evaluation of instance segmentation vs. patch-based classification for honeycomb detection}
Models detecting honeycombs by classifying patches and by instance segmentation were trained on \gls{hicc} and \gls{hicis} respectively.
EfficientNet-B0 trained on \gls{hicc} \emph{metis-s112-p224} achieved the overall highest performance on \emph{metis-s112-p224} with 67.6\% \gls{ap}, 67.2\% \gls{ar}, 68.5\% precision, and 55.7\% recall. While these metrics are lower compared to similar work for crack classification \cite{zhang2016road, cha2017deep, ozgenel2018performance, dorafshan2018comparison, feng2017deep}, the extent of the differences is expected, considering the size, quality and complexity of the datasets.

Although this thesis applied \acs{gradCam} for its honeycomb classifier to assist in manual verification, it also shows that it can assist in segmenting positive honeycomb patches, potentially giving better localization masks. Fan \acs{etal}\ \cite{fan2018automatic} stated that in cases of a high ratio of negative to positive pixels, a model would be likely to learn to classify each pixel as negative. Therefore, the classification enabled by the trained EfficientNet-B0 model is essential for further research in this direction.

The \gls{mrcnn} trained on the \gls{hicis} \emph{web} and \emph{metis} datasets achieved an 12.4\% $AP_{IoU\geq50}$, 47.7\% precision, and 34.2\% recall on the \emph{metis} test set as well as an 25.6\% $AP_{IoU\geq50}$, 64.9\% precision, and 42.1\% recall on the \emph{web} test set.
Although these metrics, especially the \glspl{ap}, are again lower compared to similar work for crack detection \cite{murao2019concrete, yin2020deep}, following the same reasoning as for honeycomb classification, the extent of the differences is expected, considering the size, quality, and complexity of the datasets.

While comparing instance segmentation and patch-based classification is challenging, when assessed quantitatively and qualitatively, the differences between the methods were not significant enough to lead to an indisputable choice. However, the former had a slight edge. Therefore, the decision between those problem types will depend on the context of possible implementation in practice, particularly on which approach integrates better with active learning in a defect documentation system.

In conclusion, the user-friendly detection of honeycombs was addressed by instance segmentation and patch-based classification. The experiments confirmed that honeycombs can be recognized by \glspl{cnn}, although the small dataset still limits the performance.

\subsection{Outlook}
While this thesis developed an initial model for detecting honeycombs, the performance would not yet suffice in practical applications. Nevertheless, models trained on either the \gls{hicc} or \gls{hicis} datasets could be used in an active learning approach integrated into defect documentation systems, enabling future research into detecting construction defects as the difficulty in obtaining labels for construction defects will persist for the myriad of existing defect types.

However, to address all defect types it will be necessary to include context information, \acs{eg} architectural plans or the \gls{bim} model, such as research into image-based construction progress monitoring will enable \cite{rho2020automated,lei2019cnn,braun2020improving}.

In conclusion, \glspl{cnn} can detect honeycombs in concrete and will enable automated defect detection assisting humans in different degrees of automation until achieving satisfactory results without human verification.